%% file: root.tex
\documentclass[letterpaper, 10 pt, conference]{ieeeconf}  
\usepackage{times}
\usepackage{epsfig}
\usepackage{graphicx}
\usepackage{amsmath}
\usepackage{amssymb}
\usepackage{hyperref}
\usepackage{booktabs}
\usepackage[font=small,labelfont=bf]{caption}
\usepackage{url}
\usepackage[table,dvipsnames]{xcolor}
\usepackage{verbatim}
\usepackage{listing}
\usepackage{pygtex}
\usepackage{newfloat}
\DeclareFloatingEnvironment[fileext=frm,placement={!ht},name=Listing]{myfloat}
\definecolor{mygray}{gray}{0.8}

\IEEEoverridecommandlockouts                              

\title{\LARGE \bf
  Balancing a CartPole System with Reinforcement Learning - A Tutorial 
}

\author{Swagat Kumar
  \thanks{* Swagat Kumar is  with the department of Computer Science, Edge Hill University,
  Ormskirk, UK L39 4QP. email: {\tt\small
  swagat.kumar@edgehill.ac.uk}} 
}

\begin{document}
	
	\maketitle
	\thispagestyle{empty}
	\pagestyle{empty}

	\begin{abstract}
     In this paper, we provide the details of implementing various
     reinforcement learning (RL) algorithms for controlling a Cart-Pole
     system. In particular, we describe various RL concepts such
     as  Q-learning, Deep Q Networks (DQN), Double DQN, Dueling
     networks, (prioritized) experience replay and show their effect on the
     learning performance.  In the process, the readers will be
     introduced to OpenAI/Gym and Keras utilities used for
     implementing the above concepts. It is observed that DQN with PER
     provides best performance among all other architectures being
     able to solve the problem within 150 episodes. 
   \end{abstract} 
	
   \section{Introduction} \label{sec:intro}
   Reinforcement Learning (RL) is a machine learning paradigm where an
   agent learns the optimal action for a given task through its
   repeated interaction with a dynamic environment that either rewards
   or punishes the agent's action. Reinforcement learning could be
   considered as a \emph{semi-supervised} learning approach where the
   supervision signal required for training the model is made available
   indirectly in the form of rewards provided by the environment.
   Reinforcement learning is more suitable for learning dynamic
   behaviour of an agent interacting with an environment rather than
   learning static mappings between two sets of input and output
   variables. Over the years, a number of reinforcement learning
   methods and architectures have been proposed with varying success.
   However, the recent success of deep learning algorithms has revived
   the field of reinforcement learning finding renewed interest among
   researchers who are now successfully applying this to solve very
   complex problems which were considered intractable earlier
   \cite{arulkumaran2017deep}. Events such as artificial agents like
   AlphaGo beating world chapmpion Lee Sedol
   \cite{borowiec2016alphago} \cite{holcomb2018overview} or IBM Watson
   winning the game of Jeopardy \cite{ferrucci2012introduction}
   \cite{markoff2011computer} has attracted worldwide attention
   towards the rise of artificial intelligence which may 
   surpass human intelligence in the near future
   \cite{kraikivski2019seeding} \cite{fang2018will}. Reinforcement
   learning is a key paradigm to build such intelligent systems which
   can learn from its experience over time. Reinforcement algorithms
   are now being increasingly applied to Robotics, healthcare,
   recommender system, data centres, smart grids, stock markets and
   transportation \cite{li2019reinforcement}. 
   
   In this paper, we will provide the implementation details of two
   well known reinforcement learning methods, namely, Q-learning
   \cite{watkins1992q} and Deep Q network (DQN) \cite{mnih2015human}
   for controlling a CartPole system. The objective is to provide a
   practical guide for implementing several reinforcement learning
   concepts by using using Python, OpenAI/Gym \cite{gym} and Keras
   \cite{gulli2017deep}. Some of these concepts are $\epsilon$-greedy
   policy, Q-learning algorithm, Deep Q-learning, experience replay,
   Dueling networks etc. It will be useful to students and researchers
   willing to venture into this field. Theoretical details and
   mathematical analysis of these concepts have omitted to maintain
   the brevity of this paper. Readers are, instead, referred to
   the relevant literature for more in-depth understanding of these
   concepts. 

   The rest of this paper is organized as follows. Various deep
   learning concepts with their implementation details are provided in
   next section. The results of applying these concepts to solve the
   CartPole problem is discussed in Section \ref{sec:expt}. The
   conclusion is provided in Section \ref{sec:conc}. 

\begin{figure}[!t]
  \centering
  \begin{tabular}{cc}
    \includegraphics[scale=0.4,trim={5cm 0 5cm 0},clip]{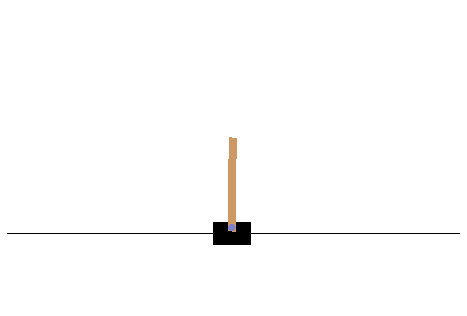}  & 
    \includegraphics[scale=0.9,trim={3cm 0 2cm 1.75cm},clip]{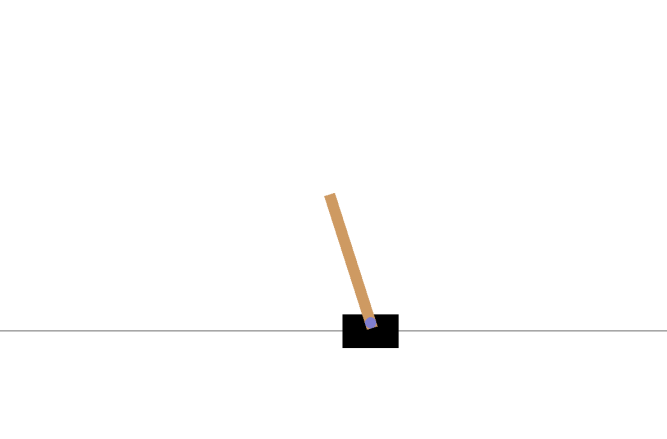}  \\
    \small{(a)} & \small{(b)}
\end{tabular}
  \caption{A Cart-Pole System: (a) Balanced state, (b) Unbalanced state}
  \label{fig:cartpole}
\end{figure}

   \section{Methods} \label{sec:meth}

   \subsection{The System}
   We use OpenAI Gym \cite{gym} to simulate the Cart-Pole system. Few
   snapshots of Cart-Pole states are shown in Figure
   \ref{fig:cartpole}. The left image shows the balanced state while
   the right image shows an imbalanced state.  It consists of a cart
   (shown in black color) and a vertical bar attached to the cart
   using passive pivot joint.  The cart can move left or right. The
   problem is to prevent the vertical bar from falling by moving the
   car left or right. One can see the animation of system behaviour
   under random action policy by executing the code given in Listing
   \ref{list:code1}. The state vector for this system $\mathbf{x}$ is
   a four dimensional vector having components $\{x,\dot{x}, \theta,
   \dot{\theta}\}$. The action has two states: left (0) and right (1).
   The episode terminates if (1) the pole angle is more than $\pm
   12^{\circ}$ from the vertical axis, or (2) the cart position is
   more than $\pm 2.4$ cm from the centre, or (3) the episode length
   is greater than 200. The agent receives a reward of 1 for every
   step taken including the termination step.  The problem is
   considered solved, if the average reward is greater than or equal
   to 195 over 100 consecutive episodes.

\begin{listing}
   \input{./pythoncode/cartpole-py}
     \caption{Simple Code to visualize Cartpole Animation}
     \label{list:code1}
\end{listing}

\subsection{Q-Learning Algorithm} \label{sec:qlearn}
Q-learning algorithm uses Bellman Equation to form a Q-function to
quantify the expected discounted future rewards that can be obtained
by taking an action $a_t$ for a given state $s_t$ at any time t. Mathematically, it can be written as: 
\begin{equation}
  Q^{\pi}(s_t,a_t) = E[R_{t+1} + \gamma R_{t+2} + \gamma^2 R_{t+3} + \cdots |(s_t, a_t)]
  \label{q-func}
\end{equation}
where $R_i,\ i = t+1,\ldots$ is the future rewards and $\gamma$ is the
discount factor. The objective is to update these Q function values
through an iterative process by exploring all possible combinations of
state and actions. Q-learning assumes a discrete observation space.
Hence, the continuous state values are first discretized into fixed
number of buckets by using {\tt bucketize()} function as shown below: 

\begin{listing}
  \input{./pythoncode/bucketize-py.tex}
  \caption{Discretizing continuous states into discrete states}
  \label{lst:bucket}
\end{listing}

The method involves creating a Q-table that stores rewards for all
possible combinations of state and action choices. With a bucket size
$(1,1,6,3)$ for states and two dimensional action vector, the
dimension of Q-table is $1\times 1 \times 6 \times 3 \times 2$. The
Q-learning algorithm is shown in Listing \ref{lst:qlearn}. It
consists of the following four major steps:

 \begin{enumerate}
   \item Select an action as per the $\epsilon$-greedy policy where
     $\epsilon$ controls the balance between exploration and
     exploitation. A random action is selected during exploration.
     During exploitation however, an action is selected based on agent's
     past experience. This is achieved by selecting an action that has
     maximum reward in the Q-table for the current state.
     Mathematically, we can write
     \begin{equation}
       a(s) = \arg\max_{a'}Q(s,a')
       \label{eq:action_exploit}
     \end{equation}
     The exploration rate $\epsilon$ starts with a value of 1.0 at the
     beginning of the training and is reduced gradually over time.
     The corresponding code for selecting action is shown below:
     \begin{listing}
       \input{./pythoncode/select_action-py}
     \end{listing}
     
   \item Obtain new observations with the above action and collect
     reward from the environment.

   \item Update the Q-table using the following formulation: 
     \begin{equation}
       Q(s,a) = Q(s,a) + \alpha [R(s,a) + \gamma \max_a Q(s',a) - Q(s,a)]
       \label{eq:q-update}
     \end{equation}
     where $\alpha$ is the learning rate which is reduce monotonically
     from 1.0 to 0.1 as the training progresses. 

   \item Update the current state and repeat the above steps in the
     the next iteration. 
 \end{enumerate}

 The initial configurations and user-defined parameters for Q-learning
 algorithm is shown in Code Listing \ref{lst:qlconf}. The actual steps
 involved in the update of Q-table during each iteration is provided
 in Code Listing \ref{lst:qlearn}. These two parts could be executed
 together as a single python program.

\begin{listing}
  \input{./pythoncode/qrlcode1-py.tex}
  \caption{Initial Configurations for Q-learning algorithm}
  \label{lst:qlconf}
\end{listing}

\begin{listing}
  \input{./pythoncode/qrlcode2-py}
  \caption{Q-learning algorithm}
  \label{lst:qlearn}
\end{listing}

 \subsection{Deep Q Network (DQN) Algorithm} \label{sec:dqn}
 Q-learning algorithm suffers from the \emph{Curse-of-Dimensionality}
 problem as it requires discrete states to form the Q-table. The
 computational complexity of Q-learning increases exponentially with
 increasing dimension of the state and action vector. Deep Q learning
 solves this problem by approximating the Q-value function $Q(s,a)$ with an
 artificial neural network. This is achieved by the function {\tt
 build\_model()} that uses Keras APIs to build a deep Q-network as shown below: 
 \begin{listing}
   \input{./pythoncode/dqn_model-py}
   \caption{Creating a DQN using Keras APIs}
   \label{lst:dqn_model}
 \end{listing}
It is a 4-24-24-2 feed-forward network with 4 inputs, 2 outputs and
 two hidden layers each having 24 nodes. Hidden nodes use a RELU
 activation function while the output layer nodes use a linear activation
 function. Having a deep network to estimate Q values allows us to
 work directly with continuous state and action values.  The Q network
 needs to be trained to estimate Q-values for a given state and action pair. This is
 done by using the following loss function: \[L_i(\theta_i) =
   E_{(s,a)\sim P(s,a)}[Q^*(s,a) - Q(s,a;\theta_i)]^2 \] where the target
   Q value $Q^*(s,a)$ for each iteration is given by 
   \begin{equation}
     Q^*(s,a) = E_{s'\in S}
     [R(s,a) + \gamma \max_{a'}Q(s',a';\theta_{i-1})|s,a]
     \label{eq:q_target}
   \end{equation}
     where $R(s,a)$ is the reward for the current state-action pair
     $(s,a)$ obtained from the environment and $Q(s',a',\theta_{i-1})$
     is the Q-value for the next state obtained using the Q-network
     weights from the last iteration. This is implemented using the
     code provided in the code listing \ref{lst:dqn_target}. It also
     shows the code for computing Q targets for DDQN architecture
     which will be explained later in this paper. 

     \begin{listing}[H]
       \input{./pythoncode/dqn_target_model-py}
       \caption{Obtaining the target Q values required for training DQN and DDQN}
       \label{lst:dqn_target}
     \end{listing}

Sometimes it is convenient to have a separate network to obtain target
Q values. It is called a target Q network $Q'$ having same
architecture as that of the original Q network. The weights for the
target network is copied from the original network at regular
intervals. This is shown in the code listing \ref{lst:wtupdate} where
the {\tt ddqn} flag needs to be set to {\tt false}.                      

\begin{listing}
  \input{./pythoncode/update_target-py}
  \caption{Weight update for the target network at regular
    intervals. Polyak Averaging can be implementing by setting ddqn
  flag.}
  \label{lst:wtupdate}
\end{listing}

     \subsection{Experience Replay}\label{sec:replay}
     It has been shown that the network trains faster with a batch
     update rather than with an incremental weight update method. In a
     batch update, the network weights are updated after applying a
     number of samples to the network whereas in incremental update,
     the network is updated after applying each sample to the network.
     In this context, DQN uses a concept called \emph{experience
     replay} where a random sample of past experiences of the agent is
     used for training the Q network. The experiences are stored in a
     fixed size replay memory in the form of tuples $(s,a,r,s')$
     containing current state, current action,
     reward and next state after each iteration. Once a sufficient
     number of entries are stored in the replay memory, we can train
     the DQN by using a batch of samples selected randomly from the
     replay memory. The exploration rate $\epsilon$ is reduced
     monotonically after each iteration of training. 

     \begin{listing}
       \input{./pythoncode/exp_replay-py}
       \caption{Training DQN using Experience Replay}
       \label{lst:expreplay}
     \end{listing}

   \subsection{Double DQN} \label{sec:ddqn}
   Taking the maximum of estimated Q value as the target value for
   training a DQN as per equation \ref{eq:q_target} may introduce a
   maximization bias in learning.  Since Q learning involves
   \emph{bootstrapping}, i.e., learning estimates from estimates, such
   overestimation may become problematic over time. This can be solved
   by using double Q learning \cite{hasselt2010double}
   \cite{van2016deep} which uses two Q-value estimators, each of which
   is used to update the other. In this paper, we implement the
   version proposed in \cite{van2016deep} that uses two models $Q$ and
   $Q'$ sharing weights at regular intervals. The network $Q'$ is used
   for action selection while the network $Q$ is used for action
   evaluation. That is, the target value for network training is
   obtained by using the following equation:
   \begin{equation}
     Q^*(s,a) \approx r_t + \gamma Q(s_{t+1},\arg\max_a'Q'(s_t,a_t))
     \label{eq:ddqn_target}
   \end{equation}
   We minimize the error between $Q$ and $Q^*$, but have $Q'$ slowly
   copy the parameters of Q through Polyak averaging: $\theta' = \tau
   \theta + (1-\tau) \theta'$. The code for computing target Q value
   and weight update is shown in code listings \ref{lst:dqn_target}
   and \ref{lst:wtupdate} respectively where the {\tt ddqn} flag
   needs to be set to {\tt true}.

   \subsection{Dueling DQN} \label{dueling}
   The Q-value $Q(s,a)$ tells us how good it is to take an action $a$
   being at state $s$. This Q-value can be decomposed as the sum of
   $V(s)$, the value of being at that state, and $A(s,a)$, the
   advantage of taking that action at the state (from all other
   possible actions). Mathematically, we can write this as 
   \begin{equation}
     Q(s,a) = V(s) + A(s,a)
     \label{eq:ddqn_qvalue}
   \end{equation}
   Dueling DQN uses two separate estimators for these two components
   which are then combined together through a special aggregation
   layer to get an estimate of $Q(s,a)$. By decoupling the estimation,
   intuitively the Dueling DQN can learn which states are (or are not)
   valuable without having to learn the effect of each action at each
   state. This is particularly useful for states where actions do not
   affect the environment in a meaningful way. In these cases, it is
   unnecessary to evaluate each action for such states and could be
   skipped to speed up the learning process. 

   Rather than directly adding individual components as shown in
   \eqref{eq:ddqn_qvalue}, the q-value estimate can be obtained by
   using the following two forms of aggregation: 
   \begin{eqnarray}
     Q(s,a) =  V(s,\beta) + A(s,a,\alpha) - \max_{a'} A(s,a',\alpha) \label{eq:qaggr_max}\\
     Q(s,a) =  V(s, \beta) + A(s,a, \alpha) - \frac{1}{|A|}\sum_{a'} A(s,a',\alpha)
     \label{eq:qaggr_avg}
   \end{eqnarray}
   where $\beta$ and $\alpha$ are the weights for the networks $V(s)$
   and $A(s,a)$ respectively. The first equation 
   \eqref{eq:qaggr_max} uses max advantage value and the second
   equation \eqref{eq:qaggr_avg} uses the average advantage value to
   estimate $Q(s,a)$ from $V(s)$.  This form of aggregation apparently
   solves the \emph{issue of identifiability}, that is - given
   $Q(s,a)$, it is difficult to find $A(s,a)$ and $V(s)$.

\begin{figure}[!h]
  \centering
  \includegraphics[scale=0.4]{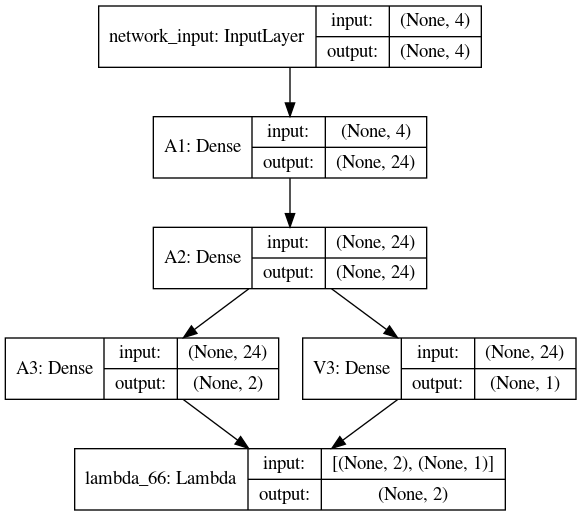}
  \caption{Dueling Model architecture for DQN}
  \label{fig:duel_model}
\end{figure}

   The implementation of Dueling DQN architecture involves replacing
   the {\tt build\_model()} function provided in Code Listing
   \ref{lst:dqn_model} with the function provided in the listing
   \ref{lst:duel_arch}. A block-diagram visualization of the dueling
   architecture is shown in Figure \ref{fig:duel_model}. It uses
   Keras' Lambda function utility to implement the final aggregation
   layer. 

\begin{listing}
  \input{./pythoncode/d3qn_model-py}
  \caption{Implementing Dueling DQN Architecture with Keras}
  \label{lst:duel_arch}
\end{listing}

   \subsection{The DQN Agent} \label{dqnagent}
   The final DQN Agent class that implements both DQN, DDQN and
   Dueling versions of these architectures will appear something as
   shown in the code listing \ref{lst:dqnagent}. The implementation
   details of functions which have been discussed earlier have been
   omitted here. Please remember to change your {\tt build\_model()}
   if you are implementing a dueling architecture. The main body of
   the program that uses this {\tt DQNAgent} class to control the
   cart-pole system is provided in Code Listing \ref{lst:dqn_main}. It
   is important to set the reward to -100 when the episode ends (or
   the {\tt done} flag is set to {\tt true}). This penalizes actions
   that prematurely terminates the episode. 

   \begin{listing}
     \input{./pythoncode/dqnagent-py}
     \caption{The DQNAgent class implementation}
     \label{lst:dqnagent}
   \end{listing}

   \begin{listing}
     \input{./pythoncode/dqn_main-py}
     \caption{The main code for using DQNAgent for balancing the CartPole System.}
     \label{lst:dqn_main}
   \end{listing}

   \subsection{Prioritized Experience Replay} \label{sec:per}
   Prioritize experience replay (PER) \cite{schaul2015prioritized} is
   based on the idea that some experiences may be more important than
   others for training, but might occur less frequently. Hence, it
   will make more sense to change the sampling distribution by using a
   criterion to define the priority of each tuple of experience. PER
   will be more useful in cases where there is a big difference
   between the predicted Q-value and its TD target value, since it
   means that there is a lot to learn about it. The priority of an
   experience is therefore defined as: \begin{equation} p_t =
     |\delta_t| + e \label{eq:priority} \end{equation} where
   $|\delta_t|$ is the magnitude of TD error and $e$ is a constant
   that ensures that no experience has zero probability of getting
   selected. Hence, the experiences are stored in the replay memory
   along with their priorities as a tuple $<s_t, a_t, r_t, s_{t+1},
   p_t>$. However, one can not simply do a greedy prioritization as it
   will lead to training with the same experiences (having bigger
   priority) and hence over-fitting. Hence, this priority is converted
   into stochastic probability given by \begin{equation} P(i) =
     \frac{p_i^a}{\sum_k p_k^a} \label{eq:priorprob} \end{equation}
   where $a$ is a hyperparameter used to reintroduce some randomness
   in the experience selection for the replay buffer. $a=0$ will lead
   to pure uniform randomness while $a=1$ will select the experiences
   with highest priorities. Priority sampling, in general, will
   introduce a bias towards high-priority samples and may lead to
   over-fitting. To correct this bias, we use important sampling (S)
   weights that will adjust the updating by reducing the weights of
   the often seen samples. The weights for each sample is given by:
   \begin{equation} w_i = \left(\frac{1}{N}.\frac{1}{P(i)}\right)^b
     \label{eq:iswt} \end{equation} The role of hyperparameter $b$ is
   to control how much these importance sampling weights affect the
   learning. In practice, $b$ is selected to be 0 in the beginning and
   is annealed upto 1 over the duration of training, because these
   weights are more important in the end of learning when the Q-values
   begin to converge. To reduce the computational burden a {\tt
   sumTree} data structure is used which provides $O(\log n)$ time
   complexity for sampling experiences and updating their priorities.
   A {\tt sumTree} is a Binary Tree, that is a tree with a maximum of
   two children for each node.  The leaves contain the priority values
   and a data array containing experiences. The code for creating {\tt
   sumTree} data structure and the corresponding replay memory is
   shown in the code block listing \ref{lst:sumtree} and
   \ref{lst:stmem} respectively. The training function {\tt
   experience\_replay()} will be slightly different from the one given
   in code listing \ref{lst:expreplay} as it will make use of sumtree
   data structure and require updating priorities with each iteration.
   The code for the modified version of {\tt experience\_replay()}
   function is provided in code listing \ref{lst:per_train}. The main
   changes in the {\tt DQNAgent} class definition is shown in the code
   listing \ref{lst:per_dqnagent}. The main program required for
   solving the Cartpole problem remains the same as given in code
   listing \ref{lst:dqn_main}. The effect of PER is discussed later in
   the experiment section. 

   \begin{listing}
     \input{./pythoncode/sumtree-py} 
     \caption{The sum tree data structure for creating replay memory.}
     \label{lst:sumtree}
   \end{listing}

   \begin{listing}
     \input{./pythoncode/st_memory-py} 
     \caption{The replay memory using sum tree data structure.}
     \label{lst:stmem}
   \end{listing}

   \begin{listing}
     \input{./pythoncode/per_train-py} 
     \caption{The code for training with prioritized experience replay.}
     \label{lst:per_train}
   \end{listing}

   \begin{listing}
     \input{./pythoncode/per_dqn_changes-py}  
     \caption{Main changes to the DQNAgent class definition provided
       in code listing \ref{lst:dqnagent}}
     \label{lst:per_dqnagent}
   \end{listing}
   
\section{Experiments and Results} \label{sec:expt} 
This section provides the details of experiments carried out to
evaluate the performance of various reinforcement learning models
described in the previous sections. This is described next in the
following subsections.

\subsection{Software and Hardware Setup}
The complete implementation code for this paper is available on GitHub
\cite{rlcodesk}. The program is written using Python and Keras APIs
\cite{gulli2017deep}. It takes about a couple of hours (2-3 hours) for
running about 1000 episodes on a HP Omen laptop with a Nvidia GeForce
RTX 2060 GPU card with 6 GB of video ram.  It is also possible to make
use of freely available GPU cloud such as Google Colab \cite{colab}
\cite{bisong2019google} or Kaggle \cite{kaggle} if you don't own a GPU
machine.  

\subsection{Performance of various RL models}
\begin{figure}[!h]
  \centering
  \includegraphics[scale=0.4]{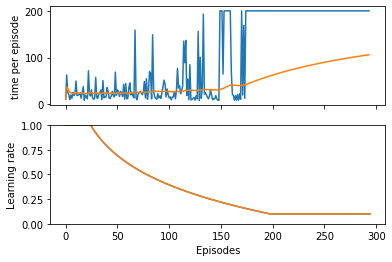}
  \caption{Performance of Q-learning Algorithm. The standard
  Q-learning solve the problem within 300 episodes. The problem is
considered solved if the average of last 100 scores is $>= 195$.
Learning rate is decreased monotonically with increasing training
episodes.}
  \label{fig:qlearn}
\end{figure}

\begin{figure}[!h]
  \centering
  \includegraphics[scale=0.4]{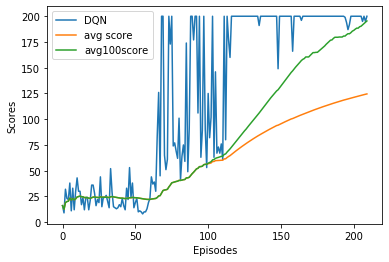}
  \caption{Performance of DQN Algorithm. Avg100score is the average of
  last 100 episodes. The problem is considered to be solved when
average of last 100 scores is $>= 195$ for CartPole-V0.}
  \label{fig:dqn}
\end{figure}

\begin{figure}[!h]
  \centering
  \includegraphics[scale=0.4]{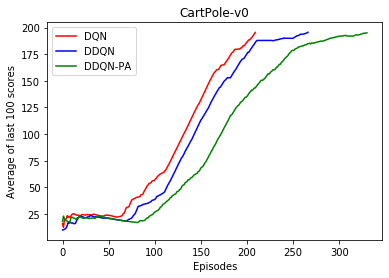}
  \caption{Performance Comparison of DQN and DDQN architectures. DQN
  is found to solve the problem faster compared to DDQN and DDQN-PA
architectures. }
  \label{fig:ddqncomp}
\end{figure}

\begin{figure}[!h]
  \centering
  \includegraphics[scale=0.4]{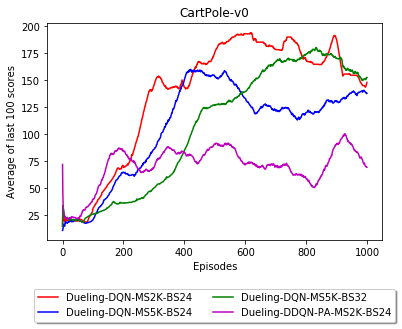}
  \caption{Performance of Dueling DQN and DDQN architectures. Dueling
  architectures fail to solve the problem within 1000 episodes.}
  \label{fig:duel-dqn}
\end{figure}

The performance of Q-learning algorithm is shown in Figure
\ref{fig:qlearn}. As one can see, Q-learning algorithm is able to
solve the problem within 300 episodes. It also shows the learning rate
that decreases monotonically with training iterations.  The
performance of DQN Algorithm with experience replay is shown in Figure
\ref{fig:dqn}. It is clearly faster than the standard Q-learning
algorithm and is found to solve the problem with 200 episodes. The
performance comparison for DQN, Double DQN (DDQN) and DDQN with Polyak
Averaging (PA) is shown in Figure \ref{fig:ddqncomp}.  While all of
them are able to solve the problem within 300 episodes, DQN is clearly
the fastest. DDQN and DDQN-PA do not provide any perceptible advantage
over DQN. This could be because the problem is itself too simple and
does not require these complex architectures.  The replay memory size
of 2000 and batch size of 24 is used for producing the result shown in
\ref{fig:ddqncomp}. Polyak Averaging (PA) tends to slow down the
learning process and it is more commonly known as the soft method for
updating target model. Similarly, the dueling versions of DQN or DDQN
architectures fail to provide convergence within 300 episodes as shown
in Figure \ref{fig:duel-dqn}.  The problem might be too simple to make
use of these complex architectures. Dueling architectures with
Prioritized Experience Replay (PER) has been shown to provide
remarkable improvement in ATARI games. It can be seen that Dueling-DQN
is faster than Dueling-DDQN as it uses less number of parameters. The
performance of DQN algorithms is also affected by changing the values
of parameters such as the replay memory size (MS) and batch size (BS)
selected for experience replay. 

\begin{table}
  \centering
  \begin{tabular}{|c|l|c|}\hline
    S.No. & Parameter & Value \\ \hline
    1  & discount factor, $\gamma$ & 0.9 \\ \hline
    2  & learning rate & 0.001 \\ \hline
    3  & Exploration rate, $\epsilon$ & 1 \\ \hline
    4  &  $\epsilon_{min}$ & 0.01 \\ \hline
    5  & polyak averaging factor, $\tau$ & 0.1 \\ \hline
  \end{tabular}
  \caption{Values of user-defined parameters used for simulation}
  \label{tab:udp}
\end{table}

\subsection{Effect of Prioritized Experience Replay}
\begin{figure}
  \centering
  \includegraphics[scale=0.4]{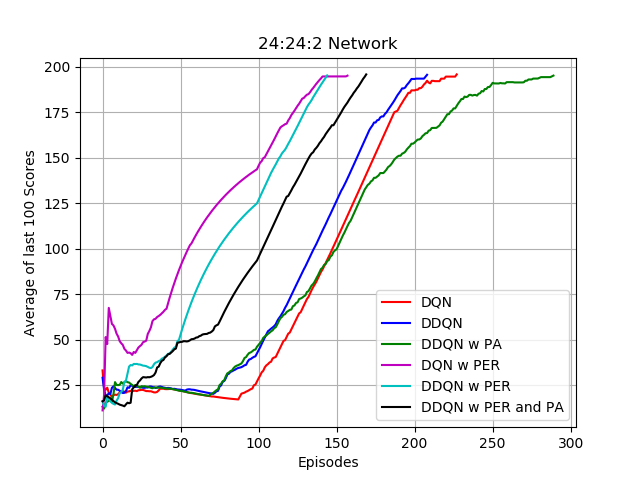} 
  \caption{Effect of Prioritized Experience Replay (PER) on DQN and
  DDQN network models. The program is terminated when the average of
last 100 scores exceed 195.}
  \label{fig:per_ddqn_result}
\end{figure}

\begin{figure}
  \centering
  \includegraphics[scale=0.4]{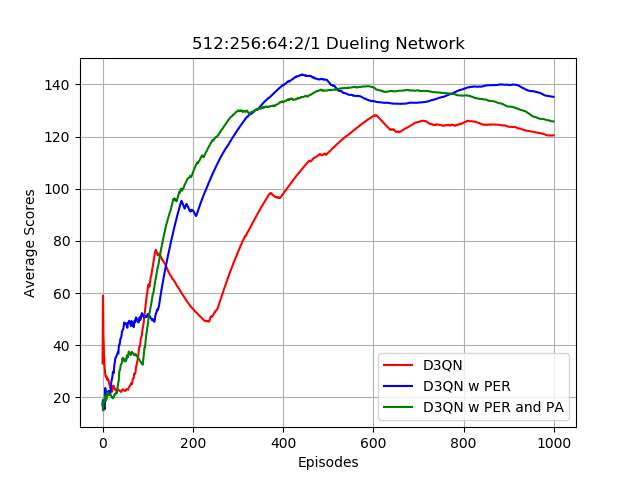}
  \caption{Effect of PER on Dueling-DDQN (D3QN) Model Architecture}
  \label{fig:per_d3qn_result}
\end{figure}

The effect of  prioritized experience replay (PER) on DQN and DDQN
architectures is shown in Figure \ref{fig:per_ddqn_result}. These
results are produced using a 3 layer network (24-24-2) architecture
with about 770 parameters, sampling batch size of 24 and a replay
memory size of 2000. The values of various hyper-parameters are as
shown in Table \ref{tab:udp}. The best performance out of 2-3
independent runs are shown in this plot. As one can see, PER provides
clear improvement over the normal DQN and DDQN implementations. The
same is seen in case of Dueling-DDQN (D3QN) architecture as shown in Figure
\ref{fig:per_d3qn_result}. This result for this figure is produced by using a
512-256-64-2/1 network architecture with about 150,531
parameters, a sampling batch size of 32 and a replay memory capacity of
10,000. As one can see, D3QN with PER provides higher average scores
compared to that obtained using only D3QN. Soft target update using
Polyak Averaging (PA) does not necessarily provide any significant
advantage over PER. It has the effect of slowing down the learning
process as is evident from these two figures. The best performance is
obtained using DQN with PER that learns to solve the problem in about
50 episodes (average of last 100 is take as the termination
criterion). 

\section{Conclusions} \label{sec:conc} 
This is a tutorial paper that provides implementation details
of a few reinforcement learning algorithms used for solving the
Cart-Pole problem. The implementation code is written in Python and
makes use of OpenAI/Gym simulation framework and Keras deep learning
tools. It is observed that DQN is considerably faster compared to the
standard Q-learning algorithms and allows the use of continuous state
values. DDQN and Dueling architectures do not provide any significant
improvement over DQN as the problem is too simple to warrant
such complex architectures. Further improvement in performance is
obtained by using Prioritized Experience Replay (PER). DQN with PER is
shown to provide the best performance so far. The codes
provided could be executed on Google Colab which provides free access
to a GPU cloud. We believe that these details will be of interest to
students and novice practitioners and will motivate them to explore
further and make novel contributions to this field. 

\section{Acknowledgement}
This work was carried out with the Refurbished Titan V
GPU provided under the NVIDIA GPU Grant 2019. The
author would like to thank NVIDIA team for their support.

\bibliographystyle{ieee} 
\bibliography{ref} 

\end{document}

%% file: pythoncode/cartpole-py.tex
\begin{Verbatim}[commandchars=\\\{\},frame=leftline,framesep=1.5ex,framerule=0.8pt,fontsize=\tiny]
\PY{k+kn}{import} \PY{n+nn}{gym}
\PY{n}{env} \PY{o}{=} \PY{n}{gym}\PY{o}{.}\PY{n}{make}\PY{p}{(}\PY{l+s+s2}{\PYZdq{}}\PY{l+s+s2}{CartPole\PYZhy{}v0}\PY{l+s+s2}{\PYZdq{}}\PY{p}{)}
\PY{n}{env}\PY{o}{.}\PY{n}{reset}\PY{p}{(}\PY{p}{)}
\PY{k}{for} \PY{n}{i\PYZus{}episode} \PY{o+ow}{in} \PY{n+nb}{range}\PY{p}{(}\PY{l+m+mi}{100}\PY{p}{)}\PY{p}{:}
  \PY{n}{obs} \PY{o}{=} \PY{n}{env}\PY{o}{.}\PY{n}{reset}\PY{p}{(}\PY{p}{)}
  \PY{n}{t} \PY{o}{=} \PY{l+m+mi}{0}
  \PY{k}{while} \PY{o+ow}{not} \PY{n}{done}\PY{p}{:}
    \PY{n}{env}\PY{o}{.}\PY{n}{render}\PY{p}{(}\PY{p}{)}
    \PY{n}{action} \PY{o}{=} \PY{n}{env}\PY{o}{.}\PY{n}{action\PYZus{}space}\PY{o}{.}\PY{n}{sample}\PY{p}{(}\PY{p}{)}
    \PY{n}{obs}\PY{p}{,}\PY{n}{reward}\PY{p}{,}\PY{n}{done}\PY{p}{,}\PY{n}{info} \PY{o}{=} \PY{n}{env}\PY{o}{.}\PY{n}{step}\PY{p}{(}\PY{n}{action}\PY{p}{)}
    \PY{n}{t} \PY{o}{+}\PY{o}{=} \PY{l+m+mi}{1}
    \PY{k}{if} \PY{n}{done}\PY{p}{:}
      \PY{k}{print}\PY{p}{(}\PY{l+s+s2}{\PYZdq{}}\PY{l+s+s2}{Done after \PYZob{}\PYZcb{} steps}\PY{l+s+s2}{\PYZdq{}}\PY{o}{.}\PY{n}{format}\PY{p}{(}\PY{n}{t}\PY{p}{)}\PY{p}{)}
      \PY{k}{break}\PY{p}{;}
\PY{n}{env}\PY{o}{.}\PY{n}{close}\PY{p}{(}\PY{p}{)}
\end{Verbatim}

%% file: pythoncode/bucketize-py.tex
\begin{Verbatim}[commandchars=\\\{\},frame=leftline,framesep=1.5ex,framerule=0.8pt,fontsize=\tiny]
\PY{k}{def} \PY{n+nf}{bucketize}\PY{p}{(}\PY{n}{state\PYZus{}value}\PY{p}{)}\PY{p}{:}
  \PY{n}{bucket\PYZus{}indices} \PY{o}{=} \PY{p}{[}\PY{p}{]}
  \PY{k}{for} \PY{n}{i} \PY{o+ow}{in} \PY{n+nb}{range}\PY{p}{(}\PY{n+nb}{len}\PY{p}{(}\PY{n}{state\PYZus{}value}\PY{p}{)}\PY{p}{)}\PY{p}{:}
    \PY{k}{if} \PY{n}{state\PYZus{}value}\PY{p}{[}\PY{n}{i}\PY{p}{]} \PY{o}{\PYZlt{}}\PY{o}{=} \PY{n}{state\PYZus{}value\PYZus{}bounds}\PY{p}{[}\PY{n}{i}\PY{p}{]}\PY{p}{[}\PY{l+m+mi}{0}\PY{p}{]}\PY{p}{:}   
        \PY{c+c1}{\PYZsh{} violates lower bound}
        \PY{n}{bucket\PYZus{}index} \PY{o}{=} \PY{l+m+mi}{0}
    \PY{k}{elif} \PY{n}{state\PYZus{}value}\PY{p}{[}\PY{n}{i}\PY{p}{]} \PY{o}{\PYZgt{}}\PY{o}{=} \PY{n}{state\PYZus{}value\PYZus{}bounds}\PY{p}{[}\PY{n}{i}\PY{p}{]}\PY{p}{[}\PY{l+m+mi}{1}\PY{p}{]}\PY{p}{:} 
        \PY{c+c1}{\PYZsh{} violates upper bound}
        \PY{c+c1}{\PYZsh{} put in the last bucket}
        \PY{n}{bucket\PYZus{}index} \PY{o}{=} \PY{n}{no\PYZus{}buckets}\PY{p}{[}\PY{n}{i}\PY{p}{]} \PY{o}{\PYZhy{}} \PY{l+m+mi}{1}  
    \PY{k}{else}\PY{p}{:}
      \PY{n}{bound\PYZus{}width} \PY{o}{=} \PY{n}{state\PYZus{}value\PYZus{}bounds}\PY{p}{[}\PY{n}{i}\PY{p}{]}\PY{p}{[}\PY{l+m+mi}{1}\PY{p}{]} \PY{o}{\PYZhy{}} \PYZbs{}
              \PY{n}{state\PYZus{}value\PYZus{}bounds}\PY{p}{[}\PY{n}{i}\PY{p}{]}\PY{p}{[}\PY{l+m+mi}{0}\PY{p}{]}
      \PY{n}{offset} \PY{o}{=} \PY{p}{(}\PY{n}{no\PYZus{}buckets}\PY{p}{[}\PY{n}{i}\PY{p}{]}\PY{o}{\PYZhy{}}\PY{l+m+mi}{1}\PY{p}{)} \PY{o}{*} \PYZbs{}
              \PY{n}{state\PYZus{}value\PYZus{}bounds}\PY{p}{[}\PY{n}{i}\PY{p}{]}\PY{p}{[}\PY{l+m+mi}{0}\PY{p}{]} \PY{o}{/} \PY{n}{bound\PYZus{}width}
      \PY{n}{scaling} \PY{o}{=} \PY{p}{(}\PY{n}{no\PYZus{}buckets}\PY{p}{[}\PY{n}{i}\PY{p}{]}\PY{o}{\PYZhy{}}\PY{l+m+mi}{1}\PY{p}{)} \PY{o}{/} \PY{n}{bound\PYZus{}width}
      \PY{n}{bucket\PYZus{}index} \PY{o}{=} \PY{n+nb}{int}\PY{p}{(}\PY{n+nb}{round}\PY{p}{(}\PY{n}{scaling}\PY{o}{*}\PY{n}{state\PYZus{}value}\PY{p}{[}\PY{n}{i}\PY{p}{]} \PY{o}{\PYZhy{}}\PY{n}{offset}\PY{p}{)}\PY{p}{)}
    \PY{n}{bucket\PYZus{}indices}\PY{o}{.}\PY{n}{append}\PY{p}{(}\PY{n}{bucket\PYZus{}index}\PY{p}{)}
  \PY{k}{return}\PY{p}{(}\PY{n+nb}{tuple}\PY{p}{(}\PY{n}{bucket\PYZus{}indices}\PY{p}{)}\PY{p}{)}
\end{Verbatim}

%% file: pythoncode/select_action-py.tex
\begin{Verbatim}[commandchars=\\\{\},frame=leftline,framesep=1.5ex,framerule=0.8pt,fontsize=\tiny]
\PY{k}{class} \PY{n+nc}{DQNAgent}\PY{p}{:}
    \PY{k}{def} \PY{n+nf}{select\PYZus{}action}\PY{p}{(}\PY{n}{state\PYZus{}value}\PY{p}{,} \PY{n}{explore\PYZus{}rate}\PY{p}{)}\PY{p}{:}
        \PY{k}{if} \PY{n}{random}\PY{o}{.}\PY{n}{random}\PY{p}{(}\PY{p}{)} \PY{o}{\PYZlt{}} \PY{n}{explore\PYZus{}rate}\PY{p}{:} 
            \PY{n}{action} \PY{o}{=} \PY{n}{env}\PY{o}{.}\PY{n}{action\PYZus{}space}\PY{o}{.}\PY{n}{sample}\PY{p}{(}\PY{p}{)}    \PY{c+c1}{\PYZsh{} explore}
        \PY{k}{else}\PY{p}{:} \PY{c+c1}{\PYZsh{} exploit}
            \PY{n}{action} \PY{o}{=} \PY{n}{np}\PY{o}{.}\PY{n}{argmax}\PY{p}{(}\PY{n}{q\PYZus{}value\PYZus{}table}\PY{p}{[}\PY{n}{state\PYZus{}value}\PY{p}{]}\PY{p}{)}  
        \PY{k}{return} \PY{n}{action}
\end{Verbatim}

%% file: pythoncode/qrlcode1-py.tex
\begin{Verbatim}[commandchars=\\\{\},frame=leftline,framesep=1.5ex,framerule=0.8pt,fontsize=\tiny]
\PY{k+kn}{import} \PY{n+nn}{gym}
\PY{k+kn}{import} \PY{n+nn}{numpy} \PY{k+kn}{as} \PY{n+nn}{np}
\PY{k+kn}{import} \PY{n+nn}{random}\PY{o}{,} \PY{n+nn}{math}
\PY{k+kn}{import} \PY{n+nn}{matplotlib.pyplot} \PY{k+kn}{as} \PY{n+nn}{plt}

\PY{n}{env}\PY{o}{=} \PY{n}{gym}\PY{o}{.}\PY{n}{make}\PY{p}{(}\PY{l+s+s1}{\PYZsq{}}\PY{l+s+s1}{CartPole\PYZhy{}v0}\PY{l+s+s1}{\PYZsq{}}\PY{p}{)}
\PY{n}{no\PYZus{}buckets} \PY{o}{=} \PY{p}{(}\PY{l+m+mi}{1}\PY{p}{,}\PY{l+m+mi}{1}\PY{p}{,}\PY{l+m+mi}{6}\PY{p}{,}\PY{l+m+mi}{3}\PY{p}{)}
\PY{n}{no\PYZus{}actions} \PY{o}{=} \PY{n}{env}\PY{o}{.}\PY{n}{action\PYZus{}space}\PY{o}{.}\PY{n}{n}
\PY{n}{state\PYZus{}value\PYZus{}bounds} \PY{o}{=} \PY{n+nb}{list}\PY{p}{(}\PY{n+nb}{zip}\PY{p}{(}\PY{n}{env}\PY{o}{.}\PY{n}{observation\PYZus{}space}\PY{o}{.}\PY{n}{low}\PY{p}{,} \PY{n}{env}\PY{o}{.}\PY{n}{observation\PYZus{}space}\PY{o}{.}\PY{n}{high}\PY{p}{)}\PY{p}{)}
\PY{n}{state\PYZus{}value\PYZus{}bounds}\PY{p}{[}\PY{l+m+mi}{1}\PY{p}{]} \PY{o}{=} \PY{p}{(}\PY{o}{\PYZhy{}}\PY{l+m+mf}{0.5}\PY{p}{,} \PY{l+m+mf}{0.5}\PY{p}{)}
\PY{n}{state\PYZus{}value\PYZus{}bounds}\PY{p}{[}\PY{l+m+mi}{3}\PY{p}{]} \PY{o}{=} \PY{p}{(}\PY{o}{\PYZhy{}}\PY{n}{math}\PY{o}{.}\PY{n}{radians}\PY{p}{(}\PY{l+m+mi}{50}\PY{p}{)}\PY{p}{,} \PY{n}{math}\PY{o}{.}\PY{n}{radians}\PY{p}{(}\PY{l+m+mi}{50}\PY{p}{)}\PY{p}{)}
\PY{c+c1}{\PYZsh{} define q\PYZus{}value\PYZus{}table \PYZhy{} it has a dimension of 1 x 1 x 6 x 3 x 2 }
\PY{n}{q\PYZus{}value\PYZus{}table} \PY{o}{=} \PY{n}{np}\PY{o}{.}\PY{n}{zeros}\PY{p}{(}\PY{n}{no\PYZus{}buckets} \PY{o}{+} \PY{p}{(}\PY{n}{no\PYZus{}actions}\PY{p}{,}\PY{p}{)}\PY{p}{)}
\PY{c+c1}{\PYZsh{} user\PYZhy{}defined parameters}
\PY{n}{min\PYZus{}explore\PYZus{}rate} \PY{o}{=} \PY{l+m+mf}{0.1}\PY{p}{;} \PY{n}{min\PYZus{}learning\PYZus{}rate} \PY{o}{=} \PY{l+m+mf}{0.1}\PY{p}{;} \PY{n}{max\PYZus{}episodes} \PY{o}{=} \PY{l+m+mi}{1000}
\PY{n}{max\PYZus{}time\PYZus{}steps} \PY{o}{=} \PY{l+m+mi}{250}\PY{p}{;} \PY{n}{streak\PYZus{}to\PYZus{}end} \PY{o}{=} \PY{l+m+mi}{120}\PY{p}{;} \PY{n}{solved\PYZus{}time} \PY{o}{=} \PY{l+m+mi}{199}\PY{p}{;} \PY{n}{discount} \PY{o}{=} \PY{l+m+mf}{0.99}
\PY{n}{no\PYZus{}streaks} \PY{o}{=} \PY{l+m+mi}{0}

\PY{c+c1}{\PYZsh{} Select an action using epsilon\PYZhy{}greedy policy}
\PY{k}{def} \PY{n+nf}{select\PYZus{}action}\PY{p}{(}\PY{n}{state\PYZus{}value}\PY{p}{,} \PY{n}{explore\PYZus{}rate}\PY{p}{)}\PY{p}{:} \PY{c+c1}{\PYZsh{} omitted}

\PY{c+c1}{\PYZsh{} change the exploration rate over time.}
\PY{k}{def} \PY{n+nf}{select\PYZus{}explore\PYZus{}rate}\PY{p}{(}\PY{n}{x}\PY{p}{)}\PY{p}{:}
  \PY{k}{return} \PY{n+nb}{max}\PY{p}{(}\PY{n}{min\PYZus{}learning\PYZus{}rate}\PY{p}{,} \PY{n+nb}{min}\PY{p}{(}\PY{l+m+mf}{1.0}\PY{p}{,} \PY{l+m+mf}{1.0} \PY{o}{\PYZhy{}} \PY{n}{math}\PY{o}{.}\PY{n}{log10}\PY{p}{(}\PY{p}{(}\PY{n}{x}\PY{o}{+}\PY{l+m+mi}{1}\PY{p}{)}\PY{o}{/}\PY{l+m+mi}{25}\PY{p}{)}\PY{p}{)}\PY{p}{)}

\PY{c+c1}{\PYZsh{} Change learning rate over time}
\PY{k}{def} \PY{n+nf}{select\PYZus{}learning\PYZus{}rate}\PY{p}{(}\PY{n}{x}\PY{p}{)}\PY{p}{:}
  \PY{k}{return} \PY{n+nb}{max}\PY{p}{(}\PY{n}{min\PYZus{}learning\PYZus{}rate}\PY{p}{,} \PY{n+nb}{min}\PY{p}{(}\PY{l+m+mf}{1.0}\PY{p}{,} \PY{l+m+mf}{1.0} \PY{o}{\PYZhy{}} \PY{n}{math}\PY{o}{.}\PY{n}{log10}\PY{p}{(}\PY{p}{(}\PY{n}{x}\PY{o}{+}\PY{l+m+mi}{1}\PY{p}{)}\PY{o}{/}\PY{l+m+mi}{25}\PY{p}{)}\PY{p}{)}\PY{p}{)}

\PY{c+c1}{\PYZsh{} Bucketize the state\PYZus{}value}
\PY{k}{def} \PY{n+nf}{bucketize}\PY{p}{(}\PY{n}{state\PYZus{}value}\PY{p}{)}\PY{p}{:}\PY{c+c1}{\PYZsh{} omitted}
\end{Verbatim}

%% file: pythoncode/qrlcode2-py.tex
\begin{Verbatim}[commandchars=\\\{\},frame=leftline,framesep=1.5ex,framerule=0.8pt,fontsize=\tiny]
\PY{c+c1}{\PYZsh{} train the system}
\PY{n}{totaltime} \PY{o}{=} \PY{l+m+mi}{0}
\PY{k}{for} \PY{n}{episode\PYZus{}no} \PY{o+ow}{in} \PY{n+nb}{range}\PY{p}{(}\PY{n}{max\PYZus{}episodes}\PY{p}{)}\PY{p}{:}
  \PY{c+c1}{\PYZsh{}learning rate and explore rate diminishes }
  \PY{c+c1}{\PYZsh{} monotonically  over time}
  \PY{n}{explore\PYZus{}rate} \PY{o}{=} \PY{n}{select\PYZus{}explore\PYZus{}rate}\PY{p}{(}\PY{n}{episode\PYZus{}no}\PY{p}{)}
  \PY{n}{learning\PYZus{}rate} \PY{o}{=} \PY{n}{select\PYZus{}learning\PYZus{}rate}\PY{p}{(}\PY{n}{episode\PYZus{}no}\PY{p}{)}
  \PY{c+c1}{\PYZsh{} initialize  the environment }
  \PY{n}{observation} \PY{o}{=} \PY{n}{env}\PY{o}{.}\PY{n}{reset}\PY{p}{(}\PY{p}{)}
  \PY{n}{start\PYZus{}state\PYZus{}value} \PY{o}{=} \PY{n}{bucketize\PYZus{}state\PYZus{}value}\PY{p}{(}\PY{n}{observation}\PY{p}{)}
  \PY{n}{previous\PYZus{}state\PYZus{}value} \PY{o}{=} \PY{n}{start\PYZus{}state\PYZus{}value}
  \PY{n}{done} \PY{o}{=} \PY{n+nb+bp}{False} 
  \PY{n}{time\PYZus{}step} \PY{o}{=} \PY{l+m+mi}{0}
  \PY{k}{while} \PY{o+ow}{not} \PY{n}{done}\PY{p}{:}  
    \PY{c+c1}{\PYZsh{}env.render()}
    \PY{c+c1}{\PYZsh{} select action using epsilon\PYZhy{}greedy policy}
    \PY{n}{action} \PY{o}{=} \PY{n}{select\PYZus{}action}\PY{p}{(}\PY{n}{previous\PYZus{}state\PYZus{}value}\PY{p}{,} \PY{n}{explore\PYZus{}rate}\PY{p}{)}
    \PY{c+c1}{\PYZsh{} record new observations}
    \PY{n}{observation}\PY{p}{,} \PY{n}{reward\PYZus{}gain}\PY{p}{,} \PY{n}{done}\PY{p}{,} \PY{n}{info} \PY{o}{=} \PY{n}{env}\PY{o}{.}\PY{n}{step}\PY{p}{(}\PY{n}{action}\PY{p}{)}
    \PY{c+c1}{\PYZsh{}update q\PYZus{}value\PYZus{}table}
    \PY{n}{best\PYZus{}q\PYZus{}value} \PY{o}{=} \PY{n}{np}\PY{o}{.}\PY{n}{max}\PY{p}{(}\PY{n}{q\PYZus{}value\PYZus{}table}\PY{p}{[}\PY{n}{state\PYZus{}value}\PY{p}{]}\PY{p}{)}
    \PY{n}{q\PYZus{}value\PYZus{}table}\PY{p}{[}\PY{n}{previous\PYZus{}state\PYZus{}value}\PY{p}{]}\PY{p}{[}\PY{n}{action}\PY{p}{]} \PY{o}{+}\PY{o}{=} \PY{n}{learning\PYZus{}rate} \PY{o}{*} \PY{p}{(}
        \PY{n}{reward\PYZus{}gain} \PY{o}{+} \PY{n}{discount} \PY{o}{*} \PY{n}{best\PYZus{}q\PYZus{}value} \PY{o}{\PYZhy{}} 
        \PY{n}{q\PYZus{}value\PYZus{}table}\PY{p}{[}\PY{n}{previous\PYZus{}state\PYZus{}value}\PY{p}{]}\PY{p}{[}\PY{n}{action}\PY{p}{]}\PY{p}{)}
    \PY{c+c1}{\PYZsh{} update the states for next iteration}
    \PY{n}{state\PYZus{}value} \PY{o}{=} \PY{n}{bucketize\PYZus{}state\PYZus{}value}\PY{p}{(}\PY{n}{observation}\PY{p}{)}
    \PY{n}{previous\PYZus{}state\PYZus{}value} \PY{o}{=} \PY{n}{state\PYZus{}value}
    \PY{n}{time\PYZus{}step} \PY{o}{+}\PY{o}{=} \PY{l+m+mi}{1}
    \PY{c+c1}{\PYZsh{} while loop ends here}

  \PY{k}{if} \PY{n}{time\PYZus{}step} \PY{o}{\PYZgt{}}\PY{o}{=} \PY{n}{solved\PYZus{}time}\PY{p}{:}
    \PY{n}{no\PYZus{}streaks} \PY{o}{+}\PY{o}{=} \PY{l+m+mi}{1}
  \PY{k}{else}\PY{p}{:}
    \PY{n}{no\PYZus{}streaks} \PY{o}{=} \PY{l+m+mi}{0}
  \PY{k}{if} \PY{n}{no\PYZus{}streaks} \PY{o}{\PYZgt{}} \PY{n}{streak\PYZus{}to\PYZus{}end}\PY{p}{:}
    \PY{k}{print}\PY{p}{(}\PY{l+s+s1}{\PYZsq{}}\PY{l+s+s1}{CartPole problem is solved after \PYZob{}\PYZcb{} episodes.}\PY{l+s+s1}{\PYZsq{}}\PY{p}{,} \PY{n}{episode\PYZus{}no}\PY{p}{)}
    \PY{k}{break}
\PY{n}{env}\PY{o}{.}\PY{n}{close}\PY{p}{(}\PY{p}{)}
\end{Verbatim}

%% file: pythoncode/dqn_model-py.tex
\begin{Verbatim}[commandchars=\\\{\},frame=leftline,framesep=1.5ex,framerule=0.8pt,fontsize=\tiny]
\PY{k+kn}{from} \PY{n+nn}{keras.layers} \PY{k+kn}{import} \PY{n}{Dense}
\PY{k+kn}{from} \PY{n+nn}{keras.optimizers} \PY{k+kn}{import} \PY{n}{Adam}
\PY{k+kn}{from} \PY{n+nn}{keras.models} \PY{k+kn}{import} \PY{n}{Sequential}
\PY{k}{class} \PY{n+nc}{DQNAgent}\PY{p}{:}
   \PY{c+c1}{\PYZsh{} approximate Q\PYZhy{}function with a Neural Network}
   \PY{k}{def} \PY{n+nf}{build\PYZus{}model}\PY{p}{(}\PY{n+nb+bp}{self}\PY{p}{)}\PY{p}{:}
        \PY{n}{model} \PY{o}{=} \PY{n}{Sequential}\PY{p}{(}\PY{p}{)}
        \PY{n}{model}\PY{o}{.}\PY{n}{add}\PY{p}{(}\PY{n}{Dense}\PY{p}{(}\PY{l+m+mi}{24}\PY{p}{,} \PY{n}{input\PYZus{}dim}\PY{o}{=}\PY{n+nb+bp}{self}\PY{o}{.}\PY{n}{state\PYZus{}size}\PY{p}{,} \PY{n}{activation}\PY{o}{=}\PY{l+s+s1}{\PYZsq{}}\PY{l+s+s1}{relu}\PY{l+s+s1}{\PYZsq{}}\PY{p}{)}\PY{p}{)}
        \PY{n}{model}\PY{o}{.}\PY{n}{add}\PY{p}{(}\PY{n}{Dense}\PY{p}{(}\PY{l+m+mi}{24}\PY{p}{,} \PY{n}{activation}\PY{o}{=}\PY{l+s+s1}{\PYZsq{}}\PY{l+s+s1}{relu}\PY{l+s+s1}{\PYZsq{}}\PY{p}{)}\PY{p}{)}
        \PY{n}{model}\PY{o}{.}\PY{n}{add}\PY{p}{(}\PY{n}{Dense}\PY{p}{(}\PY{n+nb+bp}{self}\PY{o}{.}\PY{n}{action\PYZus{}size}\PY{p}{,} \PY{n}{activation}\PY{o}{=}\PY{l+s+s1}{\PYZsq{}}\PY{l+s+s1}{linear}\PY{l+s+s1}{\PYZsq{}}\PY{p}{)}\PY{p}{)}
        \PY{n}{model}\PY{o}{.}\PY{n}{summary}\PY{p}{(}\PY{p}{)}
        \PY{n}{model}\PY{o}{.}\PY{n}{compile}\PY{p}{(}\PY{n}{loss}\PY{o}{=}\PY{l+s+s1}{\PYZsq{}}\PY{l+s+s1}{mse}\PY{l+s+s1}{\PYZsq{}}\PY{p}{,} \PY{n}{optimizer}\PY{o}{=}\PY{n}{Adam}\PY{p}{(}\PY{n}{lr}\PY{o}{=}\PY{n+nb+bp}{self}\PY{o}{.}\PY{n}{learning\PYZus{}rate}\PY{p}{)}\PY{p}{)}
        \PY{k}{return} \PY{n}{model}
\end{Verbatim}

%% file: pythoncode/dqn_target_model-py.tex
\begin{Verbatim}[commandchars=\\\{\},frame=leftline,framesep=1.5ex,framerule=0.8pt,fontsize=\tiny]
\PY{k}{class} \PY{n+nc}{DQNAgent}\PY{p}{:}
   \PY{k}{def} \PY{n+nf}{get\PYZus{}target\PYZus{}q\PYZus{}value}\PY{p}{(}\PY{n+nb+bp}{self}\PY{p}{,} \PY{n}{next\PYZus{}state}\PY{p}{,} \PY{n}{reward}\PY{p}{)}\PY{p}{:}
        \PY{c+c1}{\PYZsh{} max Q value among the next state\PYZsq{}s action}
        \PY{k}{if} \PY{n+nb+bp}{self}\PY{o}{.}\PY{n}{ddqn}\PY{p}{:}
            \PY{c+c1}{\PYZsh{} DDQN}
            \PY{c+c1}{\PYZsh{} Current Q network selects the action}
            \PY{c+c1}{\PYZsh{} a\PYZsq{}\PYZus{}max = argmax\PYZus{}a\PYZsq{} Q(s\PYZsq{},a\PYZsq{})}
            \PY{n}{action} \PY{o}{=} \PY{n}{np}\PY{o}{.}\PY{n}{argmax}\PY{p}{(}\PY{n+nb+bp}{self}\PY{o}{.}\PY{n}{model}\PY{o}{.}\PY{n}{predict}\PY{p}{(}\PY{n}{next\PYZus{}state}\PY{p}{)}\PY{p}{[}\PY{l+m+mi}{0}\PY{p}{]}\PY{p}{)}
            \PY{c+c1}{\PYZsh{} target Q network evaluates the action}
            \PY{c+c1}{\PYZsh{} Q\PYZus{}max = Q\PYZus{}target(s\PYZsq{}, a\PYZsq{}\PYZus{}max)}
            \PY{n}{max\PYZus{}q\PYZus{}value} \PY{o}{=} \PY{n+nb+bp}{self}\PY{o}{.}\PY{n}{target\PYZus{}model}\PY{o}{.}\PY{n}{predict}\PY{p}{(}\PY{n}{next\PYZus{}state}\PY{p}{)}\PY{p}{[}\PY{l+m+mi}{0}\PY{p}{]}\PY{p}{[}\PY{n}{action}\PY{p}{]}
        \PY{k}{else}\PY{p}{:} 
            \PY{c+c1}{\PYZsh{} DQN chooses the max Q value among next actions}
            \PY{c+c1}{\PYZsh{} Selection and evaluation of action is on the target Q network}
            \PY{c+c1}{\PYZsh{} Q\PYZus{}max = max\PYZus{}a\PYZsq{} Q\PYZus{}target(s\PYZsq{}, a\PYZsq{})}
            \PY{n}{max\PYZus{}q\PYZus{}value} \PY{o}{=} \PY{n}{np}\PY{o}{.}\PY{n}{amax}\PY{p}{(}\PY{n+nb+bp}{self}\PY{o}{.}\PY{n}{target\PYZus{}model}\PY{o}{.}\PY{n}{predict}\PY{p}{(}\PY{n}{next\PYZus{}state}\PY{p}{)}\PY{p}{[}\PY{l+m+mi}{0}\PY{p}{]}\PY{p}{)}
        \PY{k}{return} \PY{n}{max\PYZus{}q\PYZus{}value}
\end{Verbatim}

%% file: pythoncode/update_target-py.tex
\begin{Verbatim}[commandchars=\\\{\},frame=leftline,framesep=1.5ex,framerule=0.8pt,fontsize=\tiny]
\PY{k}{class} \PY{n+nc}{DQNAgent}\PY{p}{:}
    \PY{k}{def} \PY{n+nf}{update\PYZus{}target\PYZus{}model}\PY{p}{(}\PY{n+nb+bp}{self}\PY{p}{,} \PY{n}{tau} \PY{o}{=} \PY{l+m+mf}{0.1}\PY{p}{)}\PY{p}{:}
        \PY{l+s+sd}{\PYZsq{}\PYZsq{}\PYZsq{}Apply Polyak Averaging during weight update}
\PY{l+s+sd}{        make tau = 1.0 for normal update}
\PY{l+s+sd}{        \PYZsq{}\PYZsq{}\PYZsq{}}
        \PY{k}{if} \PY{n+nb+bp}{self}\PY{o}{.}\PY{n}{ddqn}\PY{p}{:} \PY{c+c1}{\PYZsh{} for DDQN}
            \PY{n}{weights} \PY{o}{=} \PY{n+nb+bp}{self}\PY{o}{.}\PY{n}{model}\PY{o}{.}\PY{n}{get\PYZus{}weights}\PY{p}{(}\PY{p}{)}
            \PY{n}{target\PYZus{}weights} \PY{o}{=} \PY{n+nb+bp}{self}\PY{o}{.}\PY{n}{target\PYZus{}model}\PY{o}{.}\PY{n}{get\PYZus{}weights}\PY{p}{(}\PY{p}{)}
            \PY{k}{for} \PY{n}{i} \PY{o+ow}{in} \PY{n+nb}{range}\PY{p}{(}\PY{n+nb}{len}\PY{p}{(}\PY{n}{target\PYZus{}weights}\PY{p}{)}\PY{p}{)}\PY{p}{:}
                \PY{n}{target\PYZus{}weights}\PY{p}{[}\PY{n}{i}\PY{p}{]} \PY{o}{=} \PY{n}{weights}\PY{p}{[}\PY{n}{i}\PY{p}{]} \PY{o}{*} \PY{n}{tau} \PY{o}{+} \PY{n}{target\PYZus{}weights}\PY{p}{[}\PY{n}{i}\PY{p}{]} \PY{o}{*} \PY{p}{(}\PY{l+m+mi}{1}\PY{o}{\PYZhy{}}\PY{n}{tau}\PY{p}{)}
            \PY{c+c1}{\PYZsh{} end of for loop}
            \PY{n+nb+bp}{self}\PY{o}{.}\PY{n}{target\PYZus{}model}\PY{o}{.}\PY{n}{set\PYZus{}weights}\PY{p}{(}\PY{n}{target\PYZus{}weights}\PY{p}{)}
        \PY{k}{else}\PY{p}{:}  \PY{c+c1}{\PYZsh{} for DQN}
            \PY{n+nb+bp}{self}\PY{o}{.}\PY{n}{target\PYZus{}model}\PY{o}{.}\PY{n}{set\PYZus{}weights}\PY{p}{(}\PY{n+nb+bp}{self}\PY{o}{.}\PY{n}{model}\PY{o}{.}\PY{n}{get\PYZus{}weights}\PY{p}{(}\PY{p}{)}\PY{p}{)}
\end{Verbatim}

%% file: pythoncode/exp_replay-py.tex
\begin{Verbatim}[commandchars=\\\{\},frame=leftline,framesep=1.5ex,framerule=0.8pt,fontsize=\tiny]
\PY{k}{class} \PY{n+nc}{DQNAgent}\PY{p}{:}
    \PY{k}{def} \PY{n+nf}{experience\PYZus{}replay}\PY{p}{(}\PY{n+nb+bp}{self}\PY{p}{)}\PY{p}{:}
        \PY{k}{if} \PY{n+nb}{len}\PY{p}{(}\PY{n+nb+bp}{self}\PY{o}{.}\PY{n}{memory}\PY{p}{)} \PY{o}{\PYZlt{}} \PY{n+nb+bp}{self}\PY{o}{.}\PY{n}{train\PYZus{}start}\PY{p}{:}
            \PY{k}{return}
        \PY{n}{batch\PYZus{}size} \PY{o}{=} \PY{n+nb}{min}\PY{p}{(}\PY{n+nb+bp}{self}\PY{o}{.}\PY{n}{batch\PYZus{}size}\PY{p}{,} \PY{n+nb}{len}\PY{p}{(}\PY{n+nb+bp}{self}\PY{o}{.}\PY{n}{memory}\PY{p}{)}\PY{p}{)}
        \PY{n}{mini\PYZus{}batch} \PY{o}{=} \PY{n}{random}\PY{o}{.}\PY{n}{sample}\PY{p}{(}\PY{n+nb+bp}{self}\PY{o}{.}\PY{n}{memory}\PY{p}{,} \PY{n}{batch\PYZus{}size}\PY{p}{)}
        \PY{n}{state\PYZus{}batch}\PY{p}{,} \PY{n}{q\PYZus{}values\PYZus{}batch} \PY{o}{=} \PY{p}{[}\PY{p}{]}\PY{p}{,} \PY{p}{[}\PY{p}{]}
        \PY{k}{for} \PY{n}{state}\PY{p}{,} \PY{n}{action}\PY{p}{,} \PY{n}{reward}\PY{p}{,} \PY{n}{next\PYZus{}state}\PY{p}{,} \PY{n}{done} \PY{o+ow}{in} \PY{n}{mini\PYZus{}batch}\PY{p}{:}
            \PY{c+c1}{\PYZsh{} q\PYZhy{}value prediction for a given state}
            \PY{n}{q\PYZus{}values\PYZus{}cs} \PY{o}{=} \PY{n+nb+bp}{self}\PY{o}{.}\PY{n}{model}\PY{o}{.}\PY{n}{predict}\PY{p}{(}\PY{n}{state}\PY{p}{)}
            \PY{c+c1}{\PYZsh{} target q\PYZhy{}value}
            \PY{n}{max\PYZus{}q\PYZus{}value\PYZus{}ns} \PY{o}{=} \PY{n+nb+bp}{self}\PY{o}{.}\PY{n}{get\PYZus{}target\PYZus{}q\PYZus{}value}\PY{p}{(}\PY{n}{next\PYZus{}state}\PY{p}{,} \PY{n}{reward}\PY{p}{)}
            \PY{c+c1}{\PYZsh{} correction on the Q value for the action used}
            \PY{k}{if} \PY{n}{done}\PY{p}{:}
                \PY{n}{q\PYZus{}values\PYZus{}cs}\PY{p}{[}\PY{l+m+mi}{0}\PY{p}{]}\PY{p}{[}\PY{n}{action}\PY{p}{]} \PY{o}{=} \PY{n}{reward} 
            \PY{k}{else}\PY{p}{:}
                \PY{n}{q\PYZus{}values\PYZus{}cs}\PY{p}{[}\PY{l+m+mi}{0}\PY{p}{]}\PY{p}{[}\PY{n}{action}\PY{p}{]} \PY{o}{=} \PY{n}{reward} \PY{o}{+} \PYZbs{}
                              \PY{n+nb+bp}{self}\PY{o}{.}\PY{n}{discount\PYZus{}factor} \PY{o}{*} \PY{n}{max\PYZus{}q\PYZus{}value\PYZus{}ns}
            \PY{n}{state\PYZus{}batch}\PY{o}{.}\PY{n}{append}\PY{p}{(}\PY{n}{state}\PY{p}{[}\PY{l+m+mi}{0}\PY{p}{]}\PY{p}{)}
            \PY{n}{q\PYZus{}values\PYZus{}batch}\PY{o}{.}\PY{n}{append}\PY{p}{(}\PY{n}{q\PYZus{}values\PYZus{}cs}\PY{p}{[}\PY{l+m+mi}{0}\PY{p}{]}\PY{p}{)}

        \PY{c+c1}{\PYZsh{} train the Q network}
        \PY{n+nb+bp}{self}\PY{o}{.}\PY{n}{model}\PY{o}{.}\PY{n}{fit}\PY{p}{(}\PY{n}{np}\PY{o}{.}\PY{n}{array}\PY{p}{(}\PY{n}{state\PYZus{}batch}\PY{p}{)}\PY{p}{,} 
                       \PY{n}{np}\PY{o}{.}\PY{n}{array}\PY{p}{(}\PY{n}{q\PYZus{}values\PYZus{}batch}\PY{p}{)}\PY{p}{,}
                       \PY{n}{batch\PYZus{}size} \PY{o}{=} \PY{n}{batch\PYZus{}size}\PY{p}{,}
                       \PY{n}{epochs} \PY{o}{=} \PY{l+m+mi}{1}\PY{p}{,} \PY{n}{verbose} \PY{o}{=} \PY{l+m+mi}{0}\PY{p}{)}
        \PY{n+nb+bp}{self}\PY{o}{.}\PY{n}{update\PYZus{}epsilon}\PY{p}{(}\PY{p}{)}
\end{Verbatim}

%% file: pythoncode/d3qn_model-py.tex
\begin{Verbatim}[commandchars=\\\{\},frame=leftline,framesep=1.5ex,framerule=0.8pt,fontsize=\tiny]
\PY{k}{class} \PY{n+nc}{DQNAgent}\PY{p}{:}
       \PY{k}{def} \PY{n+nf}{build\PYZus{}model}\PY{p}{(}\PY{n+nb+bp}{self}\PY{p}{)}\PY{p}{:}
        \PY{c+c1}{\PYZsh{} Advantage network}
        \PY{n}{network\PYZus{}input} \PY{o}{=} \PY{n}{Input}\PY{p}{(}\PY{n}{shape}\PY{o}{=}\PY{p}{(}\PY{n+nb+bp}{self}\PY{o}{.}\PY{n}{state\PYZus{}size}\PY{p}{,}\PY{p}{)}\PY{p}{,} \PY{n}{name}\PY{o}{=}\PY{l+s+s1}{\PYZsq{}}\PY{l+s+s1}{network\PYZus{}input}\PY{l+s+s1}{\PYZsq{}}\PY{p}{)}
        \PY{n}{A1} \PY{o}{=} \PY{n}{Dense}\PY{p}{(}\PY{l+m+mi}{24}\PY{p}{,} \PY{n}{activation}\PY{o}{=}\PY{l+s+s1}{\PYZsq{}}\PY{l+s+s1}{relu}\PY{l+s+s1}{\PYZsq{}}\PY{p}{,} \PY{n}{name}\PY{o}{=}\PY{l+s+s1}{\PYZsq{}}\PY{l+s+s1}{A1}\PY{l+s+s1}{\PYZsq{}}\PY{p}{)}\PY{p}{(}\PY{n}{network\PYZus{}input}\PY{p}{)}
        \PY{n}{A2} \PY{o}{=} \PY{n}{Dense}\PY{p}{(}\PY{l+m+mi}{24}\PY{p}{,} \PY{n}{activation}\PY{o}{=}\PY{l+s+s1}{\PYZsq{}}\PY{l+s+s1}{relu}\PY{l+s+s1}{\PYZsq{}}\PY{p}{,} \PY{n}{name} \PY{o}{=}\PY{l+s+s1}{\PYZsq{}}\PY{l+s+s1}{A2}\PY{l+s+s1}{\PYZsq{}}\PY{p}{)}\PY{p}{(}\PY{n}{A1}\PY{p}{)}
        \PY{n}{A3} \PY{o}{=} \PY{n}{Dense}\PY{p}{(}\PY{n+nb+bp}{self}\PY{o}{.}\PY{n}{action\PYZus{}size}\PY{p}{,} \PY{n}{activation}\PY{o}{=}\PY{l+s+s1}{\PYZsq{}}\PY{l+s+s1}{linear}\PY{l+s+s1}{\PYZsq{}}\PY{p}{,} \PY{n}{name}\PY{o}{=}\PY{l+s+s1}{\PYZsq{}}\PY{l+s+s1}{A3}\PY{l+s+s1}{\PYZsq{}}\PY{p}{)}\PY{p}{(}\PY{n}{A2}\PY{p}{)}
        \PY{c+c1}{\PYZsh{} Value network}
        \PY{n}{V3} \PY{o}{=} \PY{n}{Dense}\PY{p}{(}\PY{l+m+mi}{1}\PY{p}{,} \PY{n}{activation}\PY{o}{=}\PY{l+s+s1}{\PYZsq{}}\PY{l+s+s1}{linear}\PY{l+s+s1}{\PYZsq{}}\PY{p}{,} \PY{n}{name}\PY{o}{=}\PY{l+s+s1}{\PYZsq{}}\PY{l+s+s1}{V3}\PY{l+s+s1}{\PYZsq{}}\PY{p}{)}\PY{p}{(}\PY{n}{A2}\PY{p}{)}
        \PY{c+c1}{\PYZsh{} Final aggregation layer to compute Q(s,a)}
        \PY{k}{if} \PY{n+nb+bp}{self}\PY{o}{.}\PY{n}{dueling\PYZus{}option} \PY{o}{==} \PY{l+s+s1}{\PYZsq{}}\PY{l+s+s1}{avg}\PY{l+s+s1}{\PYZsq{}}\PY{p}{:}
            \PY{n}{network\PYZus{}output} \PY{o}{=} \PY{n}{Lambda}\PY{p}{(}\PY{k}{lambda} \PY{n}{x}\PY{p}{:} \PY{n}{x}\PY{p}{[}\PY{l+m+mi}{0}\PY{p}{]} \PY{o}{\PYZhy{}} \PY{n}{K}\PY{o}{.}\PY{n}{mean}\PY{p}{(}\PY{n}{x}\PY{p}{[}\PY{l+m+mi}{0}\PY{p}{]}\PY{p}{)} \PY{o}{+} \PY{n}{x}\PY{p}{[}\PY{l+m+mi}{1}\PY{p}{]}\PY{p}{,}\PYZbs{}
                       \PY{n}{output\PYZus{}shape}\PY{o}{=}\PY{p}{(}\PY{n+nb+bp}{self}\PY{o}{.}\PY{n}{action\PYZus{}size}\PY{p}{,}\PY{p}{)}\PY{p}{)}\PY{p}{(}\PY{p}{[}\PY{n}{A3}\PY{p}{,}\PY{n}{V3}\PY{p}{]}\PY{p}{)}
        \PY{k}{elif} \PY{n+nb+bp}{self}\PY{o}{.}\PY{n}{dueling\PYZus{}option} \PY{o}{==} \PY{l+s+s1}{\PYZsq{}}\PY{l+s+s1}{max}\PY{l+s+s1}{\PYZsq{}}\PY{p}{:}
            \PY{n}{network\PYZus{}output} \PY{o}{=} \PY{n}{Lambda}\PY{p}{(}\PY{k}{lambda} \PY{n}{x}\PY{p}{:} \PY{n}{x}\PY{p}{[}\PY{l+m+mi}{0}\PY{p}{]} \PY{o}{\PYZhy{}} \PY{n}{K}\PY{o}{.}\PY{n}{max}\PY{p}{(}\PY{n}{x}\PY{p}{[}\PY{l+m+mi}{0}\PY{p}{]}\PY{p}{)} \PY{o}{+} \PY{n}{x}\PY{p}{[}\PY{l+m+mi}{1}\PY{p}{]}\PY{p}{,}\PYZbs{}
                       \PY{n}{output\PYZus{}shape}\PY{o}{=}\PY{p}{(}\PY{n+nb+bp}{self}\PY{o}{.}\PY{n}{action\PYZus{}size}\PY{p}{,}\PY{p}{)}\PY{p}{)}\PY{p}{(}\PY{p}{[}\PY{n}{A3}\PY{p}{,}\PY{n}{V3}\PY{p}{]}\PY{p}{)}
        \PY{k}{elif} \PY{n+nb+bp}{self}\PY{o}{.}\PY{n}{dueling\PYZus{}option} \PY{o}{==} \PY{l+s+s1}{\PYZsq{}}\PY{l+s+s1}{naive}\PY{l+s+s1}{\PYZsq{}}\PY{p}{:}
            \PY{n}{network\PYZus{}output} \PY{o}{=} \PY{n}{Lambda}\PY{p}{(}\PY{k}{lambda} \PY{n}{x}\PY{p}{:} \PY{n}{x}\PY{p}{[}\PY{l+m+mi}{0}\PY{p}{]} \PY{o}{+} \PY{p}{[}\PY{l+m+mi}{1}\PY{p}{]}\PY{p}{,}\PYZbs{}
                                   \PY{n}{output\PYZus{}shape}\PY{o}{=}\PY{p}{(}\PY{n+nb+bp}{self}\PY{o}{.}\PY{n}{action\PYZus{}size}\PY{p}{,}\PY{p}{)}\PY{p}{)}\PY{p}{(}\PY{p}{[}\PY{n}{A3}\PY{p}{,}\PY{n}{V3}\PY{p}{]}\PY{p}{)}
        \PY{k}{else}\PY{p}{:}
            \PY{k}{raise} \PY{n+ne}{Exception}\PY{p}{(}\PY{l+s+s1}{\PYZsq{}}\PY{l+s+s1}{Invalid Dueling Option}\PY{l+s+s1}{\PYZsq{}}\PY{p}{)}

        \PY{n}{model} \PY{o}{=} \PY{n}{Model}\PY{p}{(}\PY{n}{network\PYZus{}input}\PY{p}{,} \PY{n}{network\PYZus{}output}\PY{p}{)}
        \PY{n}{model}\PY{o}{.}\PY{n}{compile}\PY{p}{(}\PY{n}{loss}\PY{o}{=}\PY{l+s+s1}{\PYZsq{}}\PY{l+s+s1}{mse}\PY{l+s+s1}{\PYZsq{}}\PY{p}{,} \PY{n}{optimizer}\PY{o}{=}\PY{n}{Adam}\PY{p}{(}\PY{n}{lr}\PY{o}{=}\PY{n+nb+bp}{self}\PY{o}{.}\PY{n}{learning\PYZus{}rate}\PY{p}{)}\PY{p}{)}
        \PY{n}{model}\PY{o}{.}\PY{n}{summary}\PY{p}{(}\PY{p}{)}
        \PY{n}{plot\PYZus{}model}\PY{p}{(}\PY{n}{model}\PY{p}{,} \PY{n}{to\PYZus{}file}\PY{o}{=}\PY{l+s+s1}{\PYZsq{}}\PY{l+s+s1}{model.png}\PY{l+s+s1}{\PYZsq{}}\PY{p}{,} \PY{n}{show\PYZus{}shapes}\PY{o}{=}\PY{n+nb+bp}{True}\PY{p}{,}\PYZbs{}
                                                   \PY{n}{show\PYZus{}layer\PYZus{}names}\PY{o}{=}\PY{n+nb+bp}{True}\PY{p}{)}
        \PY{k}{return} \PY{n}{model}
\end{Verbatim}

%% file: pythoncode/dqnagent-py.tex
\begin{Verbatim}[commandchars=\\\{\},frame=leftline,framesep=1.5ex,framerule=0.8pt,fontsize=\tiny]
\PY{k+kn}{import} \PY{n+nn}{numpy} \PY{k+kn}{as} \PY{n+nn}{np}
\PY{k+kn}{import} \PY{n+nn}{random}
\PY{k+kn}{from} \PY{n+nn}{collections} \PY{k+kn}{import} \PY{n}{deque}
\PY{k}{class} \PY{n+nc}{DQNAgent}\PY{p}{:}
    \PY{k}{def} \PY{n+nf+fm}{\PYZus{}\PYZus{}init\PYZus{}\PYZus{}}\PY{p}{(}\PY{n+nb+bp}{self}\PY{p}{,} \PY{n}{state\PYZus{}size}\PY{p}{,} \PY{n}{action\PYZus{}size}\PY{p}{,} \PY{n}{ddqn\PYZus{}flag}\PY{o}{=}\PY{n+nb+bp}{False}\PY{p}{)}\PY{p}{:}
        \PY{n+nb+bp}{self}\PY{o}{.}\PY{n}{state\PYZus{}size} \PY{o}{=} \PY{n}{state\PYZus{}size}
        \PY{n+nb+bp}{self}\PY{o}{.}\PY{n}{action\PYZus{}size} \PY{o}{=} \PY{n}{action\PYZus{}size}
        \PY{c+c1}{\PYZsh{} hyper parameters for DQN}
        \PY{n+nb+bp}{self}\PY{o}{.}\PY{n}{discount\PYZus{}factor} \PY{o}{=}  \PY{l+m+mf}{0.9}
        \PY{n+nb+bp}{self}\PY{o}{.}\PY{n}{learning\PYZus{}rate} \PY{o}{=} \PY{l+m+mf}{0.001}
        \PY{n+nb+bp}{self}\PY{o}{.}\PY{n}{epsilon} \PY{o}{=} \PY{l+m+mf}{1.0}        \PY{c+c1}{\PYZsh{} explore rate}
        \PY{n+nb+bp}{self}\PY{o}{.}\PY{n}{epsilon\PYZus{}decay} \PY{o}{=} \PY{l+m+mf}{0.99}
        \PY{n+nb+bp}{self}\PY{o}{.}\PY{n}{epsilon\PYZus{}min} \PY{o}{=} \PY{l+m+mf}{0.01}
        \PY{n+nb+bp}{self}\PY{o}{.}\PY{n}{batch\PYZus{}size} \PY{o}{=} \PY{l+m+mi}{24}
        \PY{n+nb+bp}{self}\PY{o}{.}\PY{n}{train\PYZus{}start} \PY{o}{=} \PY{l+m+mi}{1000}
        \PY{n+nb+bp}{self}\PY{o}{.}\PY{n}{dueling\PYZus{}option} \PY{o}{=} \PY{l+s+s1}{\PYZsq{}}\PY{l+s+s1}{avg}\PY{l+s+s1}{\PYZsq{}}
        \PY{c+c1}{\PYZsh{} create replay memory using deque}
        \PY{n+nb+bp}{self}\PY{o}{.}\PY{n}{memory} \PY{o}{=} \PY{n}{deque}\PY{p}{(}\PY{n}{maxlen}\PY{o}{=}\PY{l+m+mi}{2000}\PY{p}{)}
        \PY{c+c1}{\PYZsh{} create main model and target model}
        \PY{n+nb+bp}{self}\PY{o}{.}\PY{n}{model} \PY{o}{=} \PY{n+nb+bp}{self}\PY{o}{.}\PY{n}{build\PYZus{}model}\PY{p}{(}\PY{p}{)}
        \PY{n+nb+bp}{self}\PY{o}{.}\PY{n}{target\PYZus{}model} \PY{o}{=} \PY{n+nb+bp}{self}\PY{o}{.}\PY{n}{build\PYZus{}model}\PY{p}{(}\PY{p}{)}
        \PY{c+c1}{\PYZsh{} initialize target model}
        \PY{n+nb+bp}{self}\PY{o}{.}\PY{n}{target\PYZus{}model}\PY{o}{.}\PY{n}{set\PYZus{}weights}\PY{p}{(}\PY{n+nb+bp}{self}\PY{o}{.}\PY{n}{model}\PY{o}{.}\PY{n}{get\PYZus{}weights}\PY{p}{(}\PY{p}{)}\PY{p}{)}

    \PY{c+c1}{\PYZsh{} approximate Q\PYZhy{}function with a Neural Network}
    \PY{k}{def} \PY{n+nf}{build\PYZus{}model}\PY{p}{(}\PY{n+nb+bp}{self}\PY{p}{)}\PY{p}{:} \PY{c+c1}{\PYZsh{} omitted}

    \PY{c+c1}{\PYZsh{} update target model at regular interval to match the main model}
    \PY{k}{def} \PY{n+nf}{update\PYZus{}target\PYZus{}model}\PY{p}{(}\PY{n+nb+bp}{self}\PY{p}{)}\PY{p}{:} \PY{c+c1}{\PYZsh{} omitted}
  
    \PY{c+c1}{\PYZsh{} get action from the main model using epsilon\PYZhy{}greedy policy}
    \PY{k}{def} \PY{n+nf}{select\PYZus{}action}\PY{p}{(}\PY{n+nb+bp}{self}\PY{p}{,} \PY{n}{state}\PY{p}{)}\PY{p}{:}
        \PY{k}{if} \PY{n}{np}\PY{o}{.}\PY{n}{random}\PY{o}{.}\PY{n}{rand}\PY{p}{(}\PY{p}{)} \PY{o}{\PYZlt{}}\PY{o}{=} \PY{n+nb+bp}{self}\PY{o}{.}\PY{n}{epsilon}\PY{p}{:}
            \PY{k}{return} \PY{n}{random}\PY{o}{.}\PY{n}{randrange}\PY{p}{(}\PY{n+nb+bp}{self}\PY{o}{.}\PY{n}{action\PYZus{}size}\PY{p}{)}
        \PY{k}{else}\PY{p}{:}
            \PY{n}{q\PYZus{}value} \PY{o}{=} \PY{n+nb+bp}{self}\PY{o}{.}\PY{n}{model}\PY{o}{.}\PY{n}{predict}\PY{p}{(}\PY{n}{state}\PY{p}{)}
            \PY{k}{return} \PY{n}{np}\PY{o}{.}\PY{n}{argmax}\PY{p}{(}\PY{n}{q\PYZus{}value}\PY{p}{[}\PY{l+m+mi}{0}\PY{p}{]}\PY{p}{)}

    \PY{c+c1}{\PYZsh{} save sample \PYZlt{}s, a, r, s\PYZsq{}\PYZgt{}. into replay memory}
    \PY{k}{def} \PY{n+nf}{add\PYZus{}experience}\PY{p}{(}\PY{n+nb+bp}{self}\PY{p}{,} \PY{n}{state}\PY{p}{,} \PY{n}{action}\PY{p}{,} \PY{n}{reward}\PY{p}{,} \PY{n}{next\PYZus{}state}\PY{p}{,} \PY{n}{done}\PY{p}{)}\PY{p}{:}
        \PY{n+nb+bp}{self}\PY{o}{.}\PY{n}{memory}\PY{o}{.}\PY{n}{append}\PY{p}{(}\PY{p}{(}\PY{n}{state}\PY{p}{,}\PY{n}{action}\PY{p}{,}\PY{n}{reward}\PY{p}{,}\PY{n}{next\PYZus{}state}\PY{p}{,}\PY{n}{done}\PY{p}{)}\PY{p}{)}

    \PY{c+c1}{\PYZsh{} Compute target Q value}
    \PY{k}{def} \PY{n+nf}{get\PYZus{}target\PYZus{}q\PYZus{}value}\PY{p}{(}\PY{n+nb+bp}{self}\PY{p}{,} \PY{n}{next\PYZus{}state}\PY{p}{,} \PY{n}{reward}\PY{p}{)}\PY{p}{:}

    \PY{c+c1}{\PYZsh{} Train the model }
    \PY{k}{def} \PY{n+nf}{experience\PYZus{}replay}\PY{p}{(}\PY{n+nb+bp}{self}\PY{p}{)}\PY{p}{:}  \PY{c+c1}{\PYZsh{} omitted}

    \PY{c+c1}{\PYZsh{} decrease exploration, increase exploitation}
    \PY{k}{def} \PY{n+nf}{update\PYZus{}epsilon}\PY{p}{(}\PY{n+nb+bp}{self}\PY{p}{)}\PY{p}{:}
        \PY{k}{if} \PY{n+nb+bp}{self}\PY{o}{.}\PY{n}{epsilon} \PY{o}{\PYZgt{}} \PY{n+nb+bp}{self}\PY{o}{.}\PY{n}{epsilon\PYZus{}min}\PY{p}{:}
            \PY{n+nb+bp}{self}\PY{o}{.}\PY{n}{epsilon} \PY{o}{*}\PY{o}{=} \PY{n+nb+bp}{self}\PY{o}{.}\PY{n}{epsilon\PYZus{}decay}
\end{Verbatim}

%% file: pythoncode/dqn_main-py.tex
\begin{Verbatim}[commandchars=\\\{\},frame=leftline,framesep=1.5ex,framerule=0.8pt,fontsize=\tiny]
\PY{k}{if} \PY{n+nv+vm}{\PYZus{}\PYZus{}name\PYZus{}\PYZus{}} \PY{o}{==} \PY{l+s+s2}{\PYZdq{}}\PY{l+s+s2}{\PYZus{}\PYZus{}main\PYZus{}\PYZus{}}\PY{l+s+s2}{\PYZdq{}}\PY{p}{:}
    \PY{c+c1}{\PYZsh{} create Gym Environment}
    \PY{n}{env} \PY{o}{=} \PY{n}{gym}\PY{o}{.}\PY{n}{make}\PY{p}{(}\PY{l+s+s1}{\PYZsq{}}\PY{l+s+s1}{CartPole\PYZhy{}v0}\PY{l+s+s1}{\PYZsq{}}\PY{p}{)}
    \PY{n}{env}\PY{o}{.}\PY{n}{seed}\PY{p}{(}\PY{l+m+mi}{0}\PY{p}{)}
    \PY{n}{state\PYZus{}size} \PY{o}{=} \PY{n}{env}\PY{o}{.}\PY{n}{observation\PYZus{}space}\PY{o}{.}\PY{n}{shape}\PY{p}{[}\PY{l+m+mi}{0}\PY{p}{]}
    \PY{n}{action\PYZus{}size} \PY{o}{=} \PY{n}{env}\PY{o}{.}\PY{n}{action\PYZus{}space}\PY{o}{.}\PY{n}{n}
    \PY{c+c1}{\PYZsh{} create a DQN model}
    \PY{n}{agent} \PY{o}{=} \PY{n}{DQNAgent}\PY{p}{(}\PY{n}{state\PYZus{}size}\PY{p}{,} \PY{n}{action\PYZus{}size}\PY{p}{)}
    \PY{n}{score} \PY{o}{=} \PY{p}{[}\PY{p}{]}
    \PY{k}{for} \PY{n}{e} \PY{o+ow}{in} \PY{n+nb}{range}\PY{p}{(}\PY{n}{EPISODES}\PY{p}{)}\PY{p}{:}
        \PY{n}{done} \PY{o}{=} \PY{n+nb+bp}{False}
        \PY{n}{t} \PY{o}{=} \PY{l+m+mi}{0}
        \PY{n}{state} \PY{o}{=} \PY{n}{env}\PY{o}{.}\PY{n}{reset}\PY{p}{(}\PY{p}{)}
        \PY{n}{state} \PY{o}{=} \PY{n}{np}\PY{o}{.}\PY{n}{reshape}\PY{p}{(}\PY{n}{state}\PY{p}{,} \PY{p}{[}\PY{l+m+mi}{1}\PY{p}{,} \PY{n}{state\PYZus{}size}\PY{p}{]}\PY{p}{)}
        \PY{k}{while} \PY{o+ow}{not} \PY{n}{done}\PY{p}{:}
            \PY{n}{action} \PY{o}{=} \PY{n}{agent}\PY{o}{.}\PY{n}{get\PYZus{}action}\PY{p}{(}\PY{n}{state}\PY{p}{)}
            \PY{n}{next\PYZus{}state}\PY{p}{,} \PY{n}{reward}\PY{p}{,} \PY{n}{done}\PY{p}{,} \PY{n}{info} \PY{o}{=} \PY{n}{env}\PY{o}{.}\PY{n}{step}\PY{p}{(}\PY{n}{action}\PY{p}{)}
            \PY{n}{next\PYZus{}state} \PY{o}{=} \PY{n}{np}\PY{o}{.}\PY{n}{reshape}\PY{p}{(}\PY{n}{next\PYZus{}state}\PY{p}{,} \PY{p}{[}\PY{l+m+mi}{1}\PY{p}{,} \PY{n}{state\PYZus{}size}\PY{p}{]}\PY{p}{)}
            \PY{n}{reward} \PY{o}{=} \PY{n}{reward} \PY{k}{if} \PY{o+ow}{not} \PY{n}{done} \PY{k}{else} \PY{o}{\PYZhy{}}\PY{l+m+mi}{100} \PY{c+c1}{\PYZsh{}important step}
            \PY{c+c1}{\PYZsh{} add \PYZlt{}s,a,r,s\PYZsq{}\PYZgt{} to replay memory}
            \PY{n}{agent}\PY{o}{.}\PY{n}{append\PYZus{}sample}\PY{p}{(}\PY{n}{state}\PY{p}{,} \PY{n}{action}\PY{p}{,} \PY{n}{reward}\PY{p}{,} \PY{n}{next\PYZus{}state}\PY{p}{,} \PY{n}{done}\PY{p}{)}
            \PY{c+c1}{\PYZsh{} Train through experience replay}
            \PY{n}{agent}\PY{o}{.}\PY{n}{experience\PYZus{}replay}\PY{p}{(}\PY{p}{)}  
            \PY{n}{t} \PY{o}{+}\PY{o}{=} \PY{l+m+mi}{1}
            \PY{n}{state} \PY{o}{=} \PY{n}{next\PYZus{}state}
            \PY{k}{if} \PY{n}{done}\PY{p}{:}
                \PY{c+c1}{\PYZsh{} update target model for each episode}
                \PY{n}{agent}\PY{o}{.}\PY{n}{update\PYZus{}target\PYZus{}model}\PY{p}{(}\PY{p}{)} 
                \PY{n}{score}\PY{o}{.}\PY{n}{append}\PY{p}{(}\PY{n}{t}\PY{p}{)}
                \PY{k}{break}
                
        \PY{c+c1}{\PYZsh{} if mean score for last 100 episode bigger than 195, stop training}
        \PY{k}{if} \PY{n}{np}\PY{o}{.}\PY{n}{mean}\PY{p}{(}\PY{n}{score}\PY{p}{[}\PY{o}{\PYZhy{}}\PY{n+nb}{min}\PY{p}{(}\PY{l+m+mi}{100}\PY{p}{,} \PY{n+nb}{len}\PY{p}{(}\PY{n}{score}\PY{p}{)}\PY{p}{)}\PY{p}{:}\PY{p}{]}\PY{p}{)} \PY{o}{\PYZgt{}}\PY{o}{=} \PY{p}{(}\PY{n}{env}\PY{o}{.}\PY{n}{spec}\PY{o}{.}\PY{n}{max\PYZus{}episode\PYZus{}steps}\PY{o}{\PYZhy{}}\PY{l+m+mi}{5}\PY{p}{)}\PY{p}{:}
            \PY{k}{print}\PY{p}{(}\PY{l+s+s1}{\PYZsq{}}\PY{l+s+s1}{Problem is solved in \PYZob{}\PYZcb{} episodes.}\PY{l+s+s1}{\PYZsq{}}\PY{o}{.}\PY{n}{format}\PY{p}{(}\PY{n}{e}\PY{p}{)}\PY{p}{)}
            \PY{k}{break}
    \PY{n}{env}\PY{o}{.}\PY{n}{close}\PY{p}{(}\PY{p}{)}
\end{Verbatim}

%% file: pythoncode/sumtree-py.tex
\begin{Verbatim}[commandchars=\\\{\},frame=leftline,framesep=1.5ex,framerule=0.8pt,fontsize=\tiny]
\PY{k+kn}{import} \PY{n+nn}{numpy} \PY{k+kn}{as} \PY{n+nn}{np}
\PY{k}{class} \PY{n+nc}{SumTree}\PY{p}{(}\PY{n+nb}{object}\PY{p}{)}\PY{p}{:}
    \PY{n}{data\PYZus{}pointer} \PY{o}{=} \PY{l+m+mi}{0}
    \PY{k}{def} \PY{n+nf+fm}{\PYZus{}\PYZus{}init\PYZus{}\PYZus{}}\PY{p}{(}\PY{n+nb+bp}{self}\PY{p}{,} \PY{n}{capacity}\PY{p}{)}\PY{p}{:}
        \PY{c+c1}{\PYZsh{} Number of leaf nodes (final nodes) that contains experiences}
        \PY{n+nb+bp}{self}\PY{o}{.}\PY{n}{capacity} \PY{o}{=} \PY{n}{capacity}   \PY{c+c1}{\PYZsh{} }
        \PY{n+nb+bp}{self}\PY{o}{.}\PY{n}{tree} \PY{o}{=} \PY{n}{np}\PY{o}{.}\PY{n}{zeros}\PY{p}{(}\PY{l+m+mi}{2} \PY{o}{*} \PY{n}{capacity} \PY{o}{\PYZhy{}} \PY{l+m+mi}{1}\PY{p}{)}
        \PY{n+nb+bp}{self}\PY{o}{.}\PY{n}{data} \PY{o}{=} \PY{n}{np}\PY{o}{.}\PY{n}{zeros}\PY{p}{(}\PY{n}{capacity}\PY{p}{,} \PY{n}{dtype}\PY{o}{=}\PY{n+nb}{object}\PY{p}{)}

    \PY{k}{def} \PY{n+nf}{add}\PY{p}{(}\PY{n+nb+bp}{self}\PY{p}{,} \PY{n}{priority}\PY{p}{,} \PY{n}{data}\PY{p}{)}\PY{p}{:}
        \PY{c+c1}{\PYZsh{} Look at what index we want to put the experience}
        \PY{n}{tree\PYZus{}index} \PY{o}{=} \PY{n+nb+bp}{self}\PY{o}{.}\PY{n}{data\PYZus{}pointer} \PY{o}{+} \PY{n+nb+bp}{self}\PY{o}{.}\PY{n}{capacity} \PY{o}{\PYZhy{}} \PY{l+m+mi}{1}
        \PY{n+nb+bp}{self}\PY{o}{.}\PY{n}{data}\PY{p}{[}\PY{n+nb+bp}{self}\PY{o}{.}\PY{n}{data\PYZus{}pointer}\PY{p}{]} \PY{o}{=} \PY{n}{data} \PY{c+c1}{\PYZsh{} Update data frame }
        \PY{n+nb+bp}{self}\PY{o}{.}\PY{n}{update} \PY{p}{(}\PY{n}{tree\PYZus{}index}\PY{p}{,} \PY{n}{priority}\PY{p}{)}  \PY{c+c1}{\PYZsh{} Update the leaf }
        \PY{n+nb+bp}{self}\PY{o}{.}\PY{n}{data\PYZus{}pointer} \PY{o}{+}\PY{o}{=} \PY{l+m+mi}{1} \PY{c+c1}{\PYZsh{} Add 1 to data\PYZus{}pointer}
        \PY{k}{if} \PY{n+nb+bp}{self}\PY{o}{.}\PY{n}{data\PYZus{}pointer} \PY{o}{\PYZgt{}}\PY{o}{=} \PY{n+nb+bp}{self}\PY{o}{.}\PY{n}{capacity}\PY{p}{:}  \PY{c+c1}{\PYZsh{} If we\PYZsq{}re above the capacity}
            \PY{n+nb+bp}{self}\PY{o}{.}\PY{n}{data\PYZus{}pointer} \PY{o}{=} \PY{l+m+mi}{0} \PY{c+c1}{\PYZsh{} we go back to first index (overwrite)}

    \PY{k}{def} \PY{n+nf}{update}\PY{p}{(}\PY{n+nb+bp}{self}\PY{p}{,} \PY{n}{tree\PYZus{}index}\PY{p}{,} \PY{n}{priority}\PY{p}{)}\PY{p}{:}
        \PY{c+c1}{\PYZsh{} Change = new priority score \PYZhy{} former priority score}
        \PY{n}{change} \PY{o}{=} \PY{n}{priority} \PY{o}{\PYZhy{}} \PY{n+nb+bp}{self}\PY{o}{.}\PY{n}{tree}\PY{p}{[}\PY{n}{tree\PYZus{}index}\PY{p}{]}
        \PY{n+nb+bp}{self}\PY{o}{.}\PY{n}{tree}\PY{p}{[}\PY{n}{tree\PYZus{}index}\PY{p}{]} \PY{o}{=} \PY{n}{priority}
        \PY{k}{while} \PY{n}{tree\PYZus{}index} \PY{o}{!=} \PY{l+m+mi}{0}\PY{p}{:} \PY{c+c1}{\PYZsh{} propagate changes through the tree}
            \PY{n}{tree\PYZus{}index} \PY{o}{=} \PY{p}{(}\PY{n}{tree\PYZus{}index} \PY{o}{\PYZhy{}} \PY{l+m+mi}{1}\PY{p}{)} \PY{o}{/}\PY{o}{/} \PY{l+m+mi}{2}
            \PY{n+nb+bp}{self}\PY{o}{.}\PY{n}{tree}\PY{p}{[}\PY{n}{tree\PYZus{}index}\PY{p}{]} \PY{o}{+}\PY{o}{=} \PY{n}{change}

    \PY{k}{def} \PY{n+nf}{get\PYZus{}leaf}\PY{p}{(}\PY{n+nb+bp}{self}\PY{p}{,} \PY{n}{v}\PY{p}{)}\PY{p}{:}
        \PY{n}{parent\PYZus{}index} \PY{o}{=} \PY{l+m+mi}{0}
        \PY{k}{while} \PY{n+nb+bp}{True}\PY{p}{:}
            \PY{n}{left\PYZus{}child\PYZus{}index} \PY{o}{=} \PY{l+m+mi}{2} \PY{o}{*} \PY{n}{parent\PYZus{}index} \PY{o}{+} \PY{l+m+mi}{1}
            \PY{n}{right\PYZus{}child\PYZus{}index} \PY{o}{=} \PY{n}{left\PYZus{}child\PYZus{}index} \PY{o}{+} \PY{l+m+mi}{1}
            \PY{c+c1}{\PYZsh{} If we reach bottom, end the search}
            \PY{k}{if} \PY{n}{left\PYZus{}child\PYZus{}index} \PY{o}{\PYZgt{}}\PY{o}{=} \PY{n+nb}{len}\PY{p}{(}\PY{n+nb+bp}{self}\PY{o}{.}\PY{n}{tree}\PY{p}{)}\PY{p}{:}
                \PY{n}{leaf\PYZus{}index} \PY{o}{=} \PY{n}{parent\PYZus{}index}
                \PY{k}{break}
            \PY{k}{else}\PY{p}{:} \PY{c+c1}{\PYZsh{} downward search, always search for a higher priority node}
                \PY{k}{if} \PY{n}{v} \PY{o}{\PYZlt{}}\PY{o}{=} \PY{n+nb+bp}{self}\PY{o}{.}\PY{n}{tree}\PY{p}{[}\PY{n}{left\PYZus{}child\PYZus{}index}\PY{p}{]}\PY{p}{:}
                    \PY{n}{parent\PYZus{}index} \PY{o}{=} \PY{n}{left\PYZus{}child\PYZus{}index}
                \PY{k}{else}\PY{p}{:}
                    \PY{n}{v} \PY{o}{\PYZhy{}}\PY{o}{=} \PY{n+nb+bp}{self}\PY{o}{.}\PY{n}{tree}\PY{p}{[}\PY{n}{left\PYZus{}child\PYZus{}index}\PY{p}{]}
                    \PY{n}{parent\PYZus{}index} \PY{o}{=} \PY{n}{right\PYZus{}child\PYZus{}index}
        \PY{n}{data\PYZus{}index} \PY{o}{=} \PY{n}{leaf\PYZus{}index} \PY{o}{\PYZhy{}} \PY{n+nb+bp}{self}\PY{o}{.}\PY{n}{capacity} \PY{o}{+} \PY{l+m+mi}{1}
        \PY{k}{return} \PY{n}{leaf\PYZus{}index}\PY{p}{,} \PY{n+nb+bp}{self}\PY{o}{.}\PY{n}{tree}\PY{p}{[}\PY{n}{leaf\PYZus{}index}\PY{p}{]}\PY{p}{,} \PY{n+nb+bp}{self}\PY{o}{.}\PY{n}{data}\PY{p}{[}\PY{n}{data\PYZus{}index}\PY{p}{]}

    \PY{n+nd}{@property}
    \PY{k}{def} \PY{n+nf}{total\PYZus{}priority}\PY{p}{(}\PY{n+nb+bp}{self}\PY{p}{)}\PY{p}{:}
        \PY{k}{return} \PY{n+nb+bp}{self}\PY{o}{.}\PY{n}{tree}\PY{p}{[}\PY{l+m+mi}{0}\PY{p}{]} \PY{c+c1}{\PYZsh{} Returns the root node }
\end{Verbatim}

%% file: pythoncode/st_memory-py.tex
\begin{Verbatim}[commandchars=\\\{\},frame=leftline,framesep=1.5ex,framerule=0.8pt,fontsize=\tiny]
\PY{k+kn}{import} \PY{n+nn}{numpy} \PY{k+kn}{as} \PY{n+nn}{np}
\PY{k}{class} \PY{n+nc}{Memory}\PY{p}{(}\PY{n+nb}{object}\PY{p}{)}\PY{p}{:}  
    \PY{c+c1}{\PYZsh{} stored as ( state, action, reward, next\PYZus{}state ) in SumTree}
    \PY{n}{PER\PYZus{}e} \PY{o}{=} \PY{l+m+mf}{0.01} \PY{c+c1}{\PYZsh{} hyper parameter }
    \PY{n}{PER\PYZus{}a} \PY{o}{=} \PY{l+m+mf}{0.6}  \PY{c+c1}{\PYZsh{} hyper parameter}
    \PY{n}{PER\PYZus{}b} \PY{o}{=} \PY{l+m+mf}{0.4}  \PY{c+c1}{\PYZsh{} importance\PYZhy{}sampling, from initial value increasing to 1}
    \PY{n}{PER\PYZus{}b\PYZus{}increment\PYZus{}per\PYZus{}sampling} \PY{o}{=} \PY{l+m+mf}{0.001}
    \PY{n}{absolute\PYZus{}error\PYZus{}upper} \PY{o}{=} \PY{l+m+mf}{1.}  \PY{c+c1}{\PYZsh{} clipped abs error}

    \PY{k}{def} \PY{n+nf+fm}{\PYZus{}\PYZus{}init\PYZus{}\PYZus{}}\PY{p}{(}\PY{n+nb+bp}{self}\PY{p}{,} \PY{n}{capacity}\PY{p}{)}\PY{p}{:}
        \PY{n+nb+bp}{self}\PY{o}{.}\PY{n}{tree} \PY{o}{=} \PY{n}{SumTree}\PY{p}{(}\PY{n}{capacity}\PY{p}{)} \PY{c+c1}{\PYZsh{} Making the tree }

    \PY{k}{def} \PY{n+nf}{store}\PY{p}{(}\PY{n+nb+bp}{self}\PY{p}{,} \PY{n}{experience}\PY{p}{)}\PY{p}{:}  \PY{c+c1}{\PYZsh{} Find the max priority}
        \PY{n}{max\PYZus{}priority} \PY{o}{=} \PY{n}{np}\PY{o}{.}\PY{n}{max}\PY{p}{(}\PY{n+nb+bp}{self}\PY{o}{.}\PY{n}{tree}\PY{o}{.}\PY{n}{tree}\PY{p}{[}\PY{o}{\PYZhy{}}\PY{n+nb+bp}{self}\PY{o}{.}\PY{n}{tree}\PY{o}{.}\PY{n}{capacity}\PY{p}{:}\PY{p}{]}\PY{p}{)}
        \PY{k}{if} \PY{n}{max\PYZus{}priority} \PY{o}{==} \PY{l+m+mi}{0}\PY{p}{:}
            \PY{n}{max\PYZus{}priority} \PY{o}{=} \PY{n+nb+bp}{self}\PY{o}{.}\PY{n}{absolute\PYZus{}error\PYZus{}upper}
        \PY{n+nb+bp}{self}\PY{o}{.}\PY{n}{tree}\PY{o}{.}\PY{n}{add}\PY{p}{(}\PY{n}{max\PYZus{}priority}\PY{p}{,} \PY{n}{experience}\PY{p}{)}   

    \PY{k}{def} \PY{n+nf}{sample}\PY{p}{(}\PY{n+nb+bp}{self}\PY{p}{,} \PY{n}{n}\PY{p}{)}\PY{p}{:}
        \PY{n}{minibatch} \PY{o}{=} \PY{p}{[}\PY{p}{]}
        \PY{n}{b\PYZus{}idx} \PY{o}{=} \PY{n}{np}\PY{o}{.}\PY{n}{empty}\PY{p}{(}\PY{p}{(}\PY{n}{n}\PY{p}{,}\PY{p}{)}\PY{p}{,} \PY{n}{dtype}\PY{o}{=}\PY{n}{np}\PY{o}{.}\PY{n}{int32}\PY{p}{)}
        \PY{n}{priority\PYZus{}segment} \PY{o}{=} \PY{n+nb+bp}{self}\PY{o}{.}\PY{n}{tree}\PY{o}{.}\PY{n}{total\PYZus{}priority} \PY{o}{/} \PY{n}{n}  \PY{c+c1}{\PYZsh{} priority segment}
        \PY{k}{for} \PY{n}{i} \PY{o+ow}{in} \PY{n+nb}{range}\PY{p}{(}\PY{n}{n}\PY{p}{)}\PY{p}{:}
            \PY{c+c1}{\PYZsh{} A value is uniformly sample from each range}
            \PY{n}{a}\PY{p}{,} \PY{n}{b} \PY{o}{=} \PY{n}{priority\PYZus{}segment} \PY{o}{*} \PY{n}{i}\PY{p}{,} \PY{n}{priority\PYZus{}segment} \PY{o}{*} \PY{p}{(}\PY{n}{i} \PY{o}{+} \PY{l+m+mi}{1}\PY{p}{)}
            \PY{n}{value} \PY{o}{=} \PY{n}{np}\PY{o}{.}\PY{n}{random}\PY{o}{.}\PY{n}{uniform}\PY{p}{(}\PY{n}{a}\PY{p}{,} \PY{n}{b}\PY{p}{)}
            \PY{c+c1}{\PYZsh{} Experience that correspond to each value is retrieved}
            \PY{n}{index}\PY{p}{,} \PY{n}{priority}\PY{p}{,} \PY{n}{data} \PY{o}{=} \PY{n+nb+bp}{self}\PY{o}{.}\PY{n}{tree}\PY{o}{.}\PY{n}{get\PYZus{}leaf}\PY{p}{(}\PY{n}{value}\PY{p}{)}
            \PY{n}{b\PYZus{}idx}\PY{p}{[}\PY{n}{i}\PY{p}{]}\PY{o}{=} \PY{n}{index}
            \PY{n}{minibatch}\PY{o}{.}\PY{n}{append}\PY{p}{(}\PY{p}{[}\PY{n}{data}\PY{p}{[}\PY{l+m+mi}{0}\PY{p}{]}\PY{p}{,}\PY{n}{data}\PY{p}{[}\PY{l+m+mi}{1}\PY{p}{]}\PY{p}{,}\PY{n}{data}\PY{p}{[}\PY{l+m+mi}{2}\PY{p}{]}\PY{p}{,}\PY{n}{data}\PY{p}{[}\PY{l+m+mi}{3}\PY{p}{]}\PY{p}{,}\PY{n}{data}\PY{p}{[}\PY{l+m+mi}{4}\PY{p}{]}\PY{p}{]}\PY{p}{)}
        \PY{k}{return} \PY{n}{b\PYZus{}idx}\PY{p}{,} \PY{n}{minibatch}

    \PY{k}{def} \PY{n+nf}{batch\PYZus{}update}\PY{p}{(}\PY{n+nb+bp}{self}\PY{p}{,} \PY{n}{tree\PYZus{}idx}\PY{p}{,} \PY{n}{abs\PYZus{}errors}\PY{p}{)}\PY{p}{:}
        \PY{n}{abs\PYZus{}errors} \PY{o}{+}\PY{o}{=} \PY{n+nb+bp}{self}\PY{o}{.}\PY{n}{PER\PYZus{}e}  \PY{c+c1}{\PYZsh{} convert to abs and avoid 0}
        \PY{n}{clipped\PYZus{}errors} \PY{o}{=} \PY{n}{np}\PY{o}{.}\PY{n}{minimum}\PY{p}{(}\PY{n}{abs\PYZus{}errors}\PY{p}{,} \PY{n+nb+bp}{self}\PY{o}{.}\PY{n}{absolute\PYZus{}error\PYZus{}upper}\PY{p}{)}
        \PY{n}{ps} \PY{o}{=} \PY{n}{np}\PY{o}{.}\PY{n}{power}\PY{p}{(}\PY{n}{clipped\PYZus{}errors}\PY{p}{,} \PY{n+nb+bp}{self}\PY{o}{.}\PY{n}{PER\PYZus{}a}\PY{p}{)}
        \PY{k}{for} \PY{n}{ti}\PY{p}{,} \PY{n}{p} \PY{o+ow}{in} \PY{n+nb}{zip}\PY{p}{(}\PY{n}{tree\PYZus{}idx}\PY{p}{,} \PY{n}{ps}\PY{p}{)}\PY{p}{:}
            \PY{n+nb+bp}{self}\PY{o}{.}\PY{n}{tree}\PY{o}{.}\PY{n}{update}\PY{p}{(}\PY{n}{ti}\PY{p}{,} \PY{n}{p}\PY{p}{)}    
\end{Verbatim}

%% file: pythoncode/per_train-py.tex
\begin{Verbatim}[commandchars=\\\{\},frame=leftline,framesep=1.5ex,framerule=0.8pt,fontsize=\tiny]
\PY{k}{class} \PY{n+nc}{DQNAgent}\PY{p}{:}
    \PY{k}{def} \PY{n+nf}{experience\PYZus{}replay}\PY{p}{(}\PY{n+nb+bp}{self}\PY{p}{)}\PY{p}{:}
        \PY{l+s+sd}{\PYZsq{}\PYZsq{}\PYZsq{} Training on Mini\PYZhy{}Batch with Prioritized Experience Replay  \PYZsq{}\PYZsq{}\PYZsq{}}
        \PY{c+c1}{\PYZsh{} create a minibatch through prioritized sampling}
        \PY{n}{tree\PYZus{}idx}\PY{p}{,} \PY{n}{mini\PYZus{}batch} \PY{o}{=} \PY{n+nb+bp}{self}\PY{o}{.}\PY{n}{memory}\PY{o}{.}\PY{n}{sample}\PY{p}{(}\PY{n+nb+bp}{self}\PY{o}{.}\PY{n}{batch\PYZus{}size}\PY{p}{)}
        \PY{n}{current\PYZus{}state} \PY{o}{=} \PY{n}{np}\PY{o}{.}\PY{n}{zeros}\PY{p}{(}\PY{p}{(}\PY{n+nb+bp}{self}\PY{o}{.}\PY{n}{batch\PYZus{}size}\PY{p}{,} \PY{n+nb+bp}{self}\PY{o}{.}\PY{n}{state\PYZus{}size}\PY{p}{)}\PY{p}{)}
        \PY{n}{next\PYZus{}state} \PY{o}{=} \PY{n}{np}\PY{o}{.}\PY{n}{zeros}\PY{p}{(}\PY{p}{(}\PY{n+nb+bp}{self}\PY{o}{.}\PY{n}{batch\PYZus{}size}\PY{p}{,} \PY{n+nb+bp}{self}\PY{o}{.}\PY{n}{state\PYZus{}size}\PY{p}{)}\PY{p}{)}
        \PY{n}{qValues} \PY{o}{=} \PY{n}{np}\PY{o}{.}\PY{n}{zeros}\PY{p}{(}\PY{p}{(}\PY{n+nb+bp}{self}\PY{o}{.}\PY{n}{batch\PYZus{}size}\PY{p}{,} \PY{n+nb+bp}{self}\PY{o}{.}\PY{n}{action\PYZus{}size}\PY{p}{)}\PY{p}{)}
        \PY{c+c1}{\PYZsh{}action, reward, done = [], [], []}
        \PY{n}{action} \PY{o}{=} \PY{n}{np}\PY{o}{.}\PY{n}{zeros}\PY{p}{(}\PY{n+nb+bp}{self}\PY{o}{.}\PY{n}{batch\PYZus{}size}\PY{p}{,} \PY{n}{dtype}\PY{o}{=}\PY{n+nb}{int}\PY{p}{)}
        \PY{n}{reward} \PY{o}{=} \PY{n}{np}\PY{o}{.}\PY{n}{zeros}\PY{p}{(}\PY{n+nb+bp}{self}\PY{o}{.}\PY{n}{batch\PYZus{}size}\PY{p}{)}
        \PY{n}{done} \PY{o}{=} \PY{n}{np}\PY{o}{.}\PY{n}{zeros}\PY{p}{(}\PY{n+nb+bp}{self}\PY{o}{.}\PY{n}{batch\PYZus{}size}\PY{p}{,} \PY{n}{dtype}\PY{o}{=}\PY{n+nb}{bool}\PY{p}{)}
        \PY{k}{for} \PY{n}{i} \PY{o+ow}{in} \PY{n+nb}{range}\PY{p}{(}\PY{n+nb+bp}{self}\PY{o}{.}\PY{n}{batch\PYZus{}size}\PY{p}{)}\PY{p}{:}
            \PY{n}{current\PYZus{}state}\PY{p}{[}\PY{n}{i}\PY{p}{]} \PY{o}{=} \PY{n}{mini\PYZus{}batch}\PY{p}{[}\PY{n}{i}\PY{p}{]}\PY{p}{[}\PY{l+m+mi}{0}\PY{p}{]}   \PY{c+c1}{\PYZsh{} current\PYZus{}state}
            \PY{n}{action}\PY{p}{[}\PY{n}{i}\PY{p}{]} \PY{o}{=} \PY{n}{mini\PYZus{}batch}\PY{p}{[}\PY{n}{i}\PY{p}{]}\PY{p}{[}\PY{l+m+mi}{1}\PY{p}{]}
            \PY{n}{reward}\PY{p}{[}\PY{n}{i}\PY{p}{]} \PY{o}{=} \PY{n}{mini\PYZus{}batch}\PY{p}{[}\PY{n}{i}\PY{p}{]}\PY{p}{[}\PY{l+m+mi}{2}\PY{p}{]}
            \PY{n}{next\PYZus{}state}\PY{p}{[}\PY{n}{i}\PY{p}{]} \PY{o}{=} \PY{n}{mini\PYZus{}batch}\PY{p}{[}\PY{n}{i}\PY{p}{]}\PY{p}{[}\PY{l+m+mi}{3}\PY{p}{]}  \PY{c+c1}{\PYZsh{} next\PYZus{}state}
            \PY{n}{done}\PY{p}{[}\PY{n}{i}\PY{p}{]} \PY{o}{=} \PY{n}{mini\PYZus{}batch}\PY{p}{[}\PY{n}{i}\PY{p}{]}\PY{p}{[}\PY{l+m+mi}{4}\PY{p}{]}
            \PY{n}{qValues}\PY{p}{[}\PY{n}{i}\PY{p}{]} \PY{o}{=} \PY{n+nb+bp}{self}\PY{o}{.}\PY{n}{model}\PY{o}{.}\PY{n}{predict}\PY{p}{(}\PY{n}{current\PYZus{}state}\PY{p}{[}\PY{n}{i}\PY{p}{]}\PYZbs{}
                                            \PY{o}{.}\PY{n}{reshape}\PY{p}{(}\PY{l+m+mi}{1}\PY{p}{,}\PY{n+nb+bp}{self}\PY{o}{.}\PY{n}{state\PYZus{}size}\PY{p}{)}\PY{p}{)}\PY{p}{[}\PY{l+m+mi}{0}\PY{p}{]}
            \PY{n}{max\PYZus{}qvalue\PYZus{}ns} \PY{o}{=}
            \PY{n+nb+bp}{self}\PY{o}{.}\PY{n}{get\PYZus{}maxQvalue\PYZus{}nextstate}\PY{p}{(}\PY{n}{next\PYZus{}state}\PY{p}{[}\PY{n}{i}\PY{p}{]}\PYZbs{}
                                         \PY{o}{.}\PY{n}{reshape}\PY{p}{(}\PY{l+m+mi}{1}\PY{p}{,}\PY{n+nb+bp}{self}\PY{o}{.}\PY{n}{state\PYZus{}size}\PY{p}{)}\PY{p}{)}
            \PY{k}{if} \PY{n}{done}\PY{p}{[}\PY{n}{i}\PY{p}{]}\PY{p}{:}
                \PY{n}{qValues}\PY{p}{[}\PY{n}{i}\PY{p}{]}\PY{p}{[}\PY{n}{action}\PY{p}{[}\PY{n}{i}\PY{p}{]}\PY{p}{]} \PY{o}{=} \PY{n}{reward}\PY{p}{[}\PY{n}{i}\PY{p}{]}
            \PY{k}{else}\PY{p}{:}
                \PY{n}{qValues}\PY{p}{[}\PY{n}{i}\PY{p}{]}\PY{p}{[}\PY{n}{action}\PY{p}{[}\PY{n}{i}\PY{p}{]}\PY{p}{]} \PY{o}{=} \PY{n}{reward}\PY{p}{[}\PY{n}{i}\PY{p}{]} \PY{o}{+} \PYZbs{}
                        \PY{n+nb+bp}{self}\PY{o}{.}\PY{n}{discount\PYZus{}factor} \PY{o}{*} \PY{n}{max\PYZus{}qvalue\PYZus{}ns}
        \PY{c+c1}{\PYZsh{} update  priority in the replay memory}
        \PY{n}{target\PYZus{}old} \PY{o}{=} \PY{n}{np}\PY{o}{.}\PY{n}{array}\PY{p}{(}\PY{n+nb+bp}{self}\PY{o}{.}\PY{n}{model}\PY{o}{.}\PY{n}{predict}\PY{p}{(}\PY{n}{current\PYZus{}state}\PY{p}{)}\PY{p}{)}
        \PY{n}{target} \PY{o}{=} \PY{n}{qValues}
        \PY{n}{indices} \PY{o}{=} \PY{n}{np}\PY{o}{.}\PY{n}{arange}\PY{p}{(}\PY{n+nb+bp}{self}\PY{o}{.}\PY{n}{batch\PYZus{}size}\PY{p}{,} \PY{n}{dtype}\PY{o}{=}\PY{n}{np}\PY{o}{.}\PY{n}{int32}\PY{p}{)}
        \PY{n}{absolute\PYZus{}errors} \PY{o}{=} \PY{n}{np}\PY{o}{.}\PY{n}{abs}\PY{p}{(}\PY{n}{target\PYZus{}old}\PY{p}{[}\PY{n}{indices}\PY{p}{,} \PYZbs{}
                \PY{n}{np}\PY{o}{.}\PY{n}{array}\PY{p}{(}\PY{n}{action}\PY{p}{)}\PY{p}{]}\PY{o}{\PYZhy{}} \PY{n}{target}\PY{p}{[}\PY{n}{indices}\PY{p}{,} \PY{n}{np}\PY{o}{.}\PY{n}{array}\PY{p}{(}\PY{n}{action}\PY{p}{)}\PY{p}{]}\PY{p}{)}
        \PY{n+nb+bp}{self}\PY{o}{.}\PY{n}{memory}\PY{o}{.}\PY{n}{batch\PYZus{}update}\PY{p}{(}\PY{n}{tree\PYZus{}idx}\PY{p}{,} \PY{n}{absolute\PYZus{}errors}\PY{p}{)} 
        \PY{c+c1}{\PYZsh{} train the model}
        \PY{n+nb+bp}{self}\PY{o}{.}\PY{n}{model}\PY{o}{.}\PY{n}{fit}\PY{p}{(}\PY{n}{current\PYZus{}state}\PY{p}{,} \PY{n}{qValues}\PY{p}{,} 
                       \PY{n}{batch\PYZus{}size} \PY{o}{=} \PY{n+nb+bp}{self}\PY{o}{.}\PY{n}{batch\PYZus{}size}\PY{p}{,}
                       \PY{n}{epochs}\PY{o}{=}\PY{l+m+mi}{1}\PY{p}{,} \PY{n}{verbose}\PY{o}{=}\PY{l+m+mi}{0}\PY{p}{)}
        \PY{c+c1}{\PYZsh{} update epsilon with each training step}
        \PY{n+nb+bp}{self}\PY{o}{.}\PY{n}{update\PYZus{}epsilon}\PY{p}{(}\PY{p}{)}
\end{Verbatim}

%% file: pythoncode/per_dqn_changes-py.tex
\begin{Verbatim}[commandchars=\\\{\},frame=leftline,framesep=1.5ex,framerule=0.8pt,fontsize=\tiny]
\PY{k+kn}{from} \PY{n+nn}{sumtree} \PY{k+kn}{import} \PY{n}{sumTree} \PY{o+ow}{and} \PY{n}{Memory}
\PY{k}{class} \PY{n+nc}{DQNAgent}\PY{p}{:}
    \PY{k}{def} \PY{n+nf+fm}{\PYZus{}\PYZus{}init\PYZus{}\PYZus{}}\PY{p}{(}\PY{n+nb+bp}{self}\PY{p}{)}\PY{p}{:}
        \PY{n+nb+bp}{self}\PY{o}{.}\PY{n}{memory} \PY{o}{=} \PY{n}{Memory}\PY{p}{(}\PY{n}{memory\PYZus{}size}\PY{p}{)}

    \PY{k}{def} \PY{n+nf}{add\PYZus{}experience}\PY{p}{(}\PY{n+nb+bp}{self}\PY{p}{,} \PY{n}{state}\PY{p}{,} \PY{n}{action}\PY{p}{,} \PY{n}{reward}\PY{p}{,} \PY{n}{next\PYZus{}state}\PY{p}{,} \PY{n}{done}\PY{p}{)}\PY{p}{:}
        \PY{n}{experience} \PY{o}{=} \PY{p}{[}\PY{n}{state}\PY{p}{,} \PY{n}{action}\PY{p}{,} \PY{n}{reward}\PY{p}{,} \PY{n}{next\PYZus{}state}\PY{p}{,} \PY{n}{done}\PY{p}{]}
        \PY{n+nb+bp}{self}\PY{o}{.}\PY{n}{memory}\PY{o}{.}\PY{n}{store}\PY{p}{(}\PY{n}{experience}\PY{p}{)}

    \PY{k}{def} \PY{n+nf}{experience\PYZus{}replay}\PY{p}{(}\PY{n+nb+bp}{self}\PY{p}{)}\PY{p}{:} \PY{c+c1}{\PYZsh{} provided separately}

    \PY{c+c1}{\PYZsh{} rest of functions remain same as before}
\end{Verbatim}